\documentclass{article}
\usepackage{graphicx, subcaption, siunitx, float, hyperref, cleveref}
\usepackage[utf8]{inputenc}

\hypersetup{
    colorlinks=true,
    linkcolor=black,
    citecolor=black,
    urlcolor=blue,
}

\usepackage{natbib}
\setcitestyle{authoryear,open={(},close={)}}

\newcommand{\changeurlcolor}[1]{\hypersetup{urlcolor=#1}} 

\nocite{*}

\usepackage[accepted]{icml2019}

\icmltitlerunning{Quantifying Generalization in Reinforcement Learning}

\begin{document}

\twocolumn[
\icmltitle{Quantifying Generalization in Reinforcement Learning}

\icmlsetsymbol{equal}{*}

\begin{icmlauthorlist}
\icmlauthor{Karl Cobbe}{op}
\icmlauthor{Oleg Klimov}{op}
\icmlauthor{Chris Hesse}{op}
\icmlauthor{Taehoon Kim}{op}
\icmlauthor{John Schulman}{op}
\end{icmlauthorlist}

\icmlaffiliation{op}{OpenAI, San Francisco, CA, USA}

\icmlcorrespondingauthor{Karl Cobbe}{karl@openai.com}

\icmlkeywords{Machine Learning, Reinforcement Learning, Transfer Learning, Generalization, Procedural Generation, ICML}

\vskip 0.3in
]

\printAffiliationsAndNotice{}

\begin{abstract}
In this paper, we investigate the problem of overfitting in deep reinforcement learning. Among the most common benchmarks in RL, it is customary to use the same environments for both training and testing. This practice offers relatively little insight into an agent's ability to generalize. We address this issue by using procedurally generated environments to construct distinct training and test sets. Most notably, we introduce a new environment called CoinRun, designed as a benchmark for generalization in RL. Using CoinRun, we find that agents overfit to surprisingly large training sets. We then show that deeper convolutional architectures improve generalization, as do methods traditionally found in supervised learning, including L2 regularization, dropout, data augmentation and batch normalization.
\end{abstract}

\section{Introduction}

Generalizing between tasks remains difficult for state of the art deep reinforcement learning (RL) algorithms. Although trained agents can solve complex tasks, they struggle to transfer their experience to new environments. Agents that have mastered ten levels in a video game often fail catastrophically when first encountering the eleventh. Humans can seamlessly generalize across such similar tasks, but this ability is largely absent in RL agents. In short, agents become overly specialized to the environments encountered during training.

That RL agents are prone to overfitting is widely appreciated, yet the most common RL benchmarks still encourage training and evaluating on the same set of environments. We believe there is a need for more metrics that evaluate generalization by explicitly separating training and test environments. In the same spirit as the Sonic Benchmark \citep{gottalearnfast}, we seek to better quantify an agent’s ability to generalize.

To begin, we train agents on CoinRun, a procedurally generated environment of our own design, and we report the surprising extent to which overfitting occurs. Using this environment, we investigate how several key algorithmic and architectural decisions impact the generalization performance of trained agents.

The main contributions of this work are as follows:

\begin{enumerate}
  \item We show that the number of training environments required for good generalization is much larger than the number used by prior work on transfer in RL.
  \item We propose a generalization metric using the CoinRun environment, and we show how this metric provides a useful signal upon which to iterate.
  \item We evaluate the impact of different convolutional architectures and forms of regularization, finding that these choices can significantly improve generalization performance.
\end{enumerate}

\section{Related Work}

Our work is most directly inspired by the Sonic Benchmark \citep{gottalearnfast}, which proposes to measure generalization performance by training and testing RL agents on distinct sets of levels in the $\textit{Sonic the Hedgehog}^\text{TM}$ video game franchise. Agents may train arbitrarily long on the training set, but are permitted only 1 million timesteps at test time to perform fine-tuning. This benchmark was designed to address the problems inherent to ``training on the test set."

\cite{gen_dqn} also address this problem, accurately recognizing that conflating train and test environments has contributed to the lack of regularization in deep RL. They propose using different game modes of Atari 2600 games to measure generalization. They turn to supervised learning for inspiration, finding that both L2 regularization and dropout can help agents learn more generalizable features.

\begin{figure*}
\centering
\includegraphics[height=5.65cm]{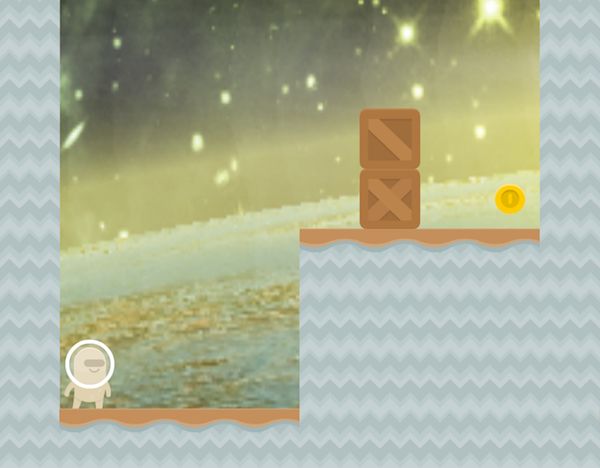}
\hspace*{\fill}
\includegraphics[height=5.65cm]{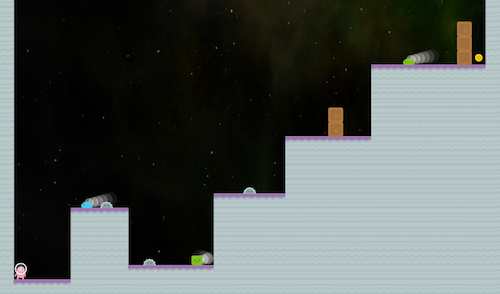}
\caption{Two levels in CoinRun. The level on the left is much easier than the level on the right.}
\label{fig:screenshots1}
\end{figure*}

\cite{assess_gen_rl} propose a different benchmark to measure generalization using six classic environments, each of which has been modified to expose several internal parameters. By training and testing on environments with different parameter ranges, their benchmark quantifies agents' ability to interpolate and extrapolate. \cite{dissect_overfitting} measure overfitting in continuous domains, finding that generalization improves as the number of training seeds increases. They also use randomized rewards to determine the extent of undesirable memorization.

Other works create distinct train and test environments using procedural generation. \cite{illuminating_gen} use the General Video Game AI (GVG-AI) framework to generate levels from several unique games. By varying difficulty settings between train and test levels, they find that RL agents regularly overfit to a particular training distribution. They further show that the ability to generalize to human-designed levels strongly depends on the level generators used during training.

\cite{overfitting} conduct experiments on procedurally generated gridworld mazes, reporting many insightful conclusions on the nature of overfitting in RL agents. They find that agents have a high capacity to memorize specific levels in a given training set, and that techniques intended to mitigate overfitting in RL, including sticky actions \citep{revisit_ale} and random starts \citep{determinism}, often fail to do so.

In \Cref{sec:env_stoch}, we similarly investigate how injecting stochasticity impacts generalization. Our work mirrors \cite{overfitting} in quantifying the relationship between overfitting and the number of training environments, though we additionally show how several methods, including some more prevalent in supervised learning, can reduce overfitting in our benchmark.

These works, as well as our own, highlight the growing need for experimental protocols that directly address generalization in RL.

\section{Quantifying Generalization} \label{sec:quant_gen}

\subsection{The CoinRun Environment}

We propose the CoinRun environment to evaluate the generalization performance of trained agents. The goal of each CoinRun level is simple: collect the single coin that lies at the end of the level. The agent controls a character that spawns on the far left, and the coin spawns on the far right. Several obstacles, both stationary and non-stationary, lie between the agent and the coin. A collision with an obstacle results in the agent’s immediate death. The only reward in the environment is obtained by collecting the coin, and this reward is a fixed positive constant. The level terminates when the agent dies, the coin is collected, or after 1000 time steps.

We designed the game CoinRun to be tractable for existing algorithms. That is, given a sufficient number of training levels and sufficient training time, our algorithms learn a near optimal policy for \textit{all} CoinRun levels. Each level is generated deterministically from a given seed, providing agents access to an arbitrarily large and easily quantifiable supply of training data. CoinRun mimics the style of platformer games like Sonic, but it is much simpler. For the purpose of evaluating generalization, this simplicity can be highly advantageous.

Levels vary widely in difficulty, so the distribution of levels naturally forms a curriculum for the agent. Two different levels are shown in \Cref{fig:screenshots1}. See \Cref{appendix:game_mech} for more details about the environment and \Cref{appendix:env_screenshots} for additional screenshots. Videos of a trained agent playing can be found \href{https://blog.openai.com/quantifying-generalization-in-reinforcement-learning}{here}, and environment code can be found \href{https://github.com/openai/coinrun}{here}.

\begin{figure*}
\centering
\begin{subfigure}{0.475 \textwidth}
\includegraphics[width=\textwidth]{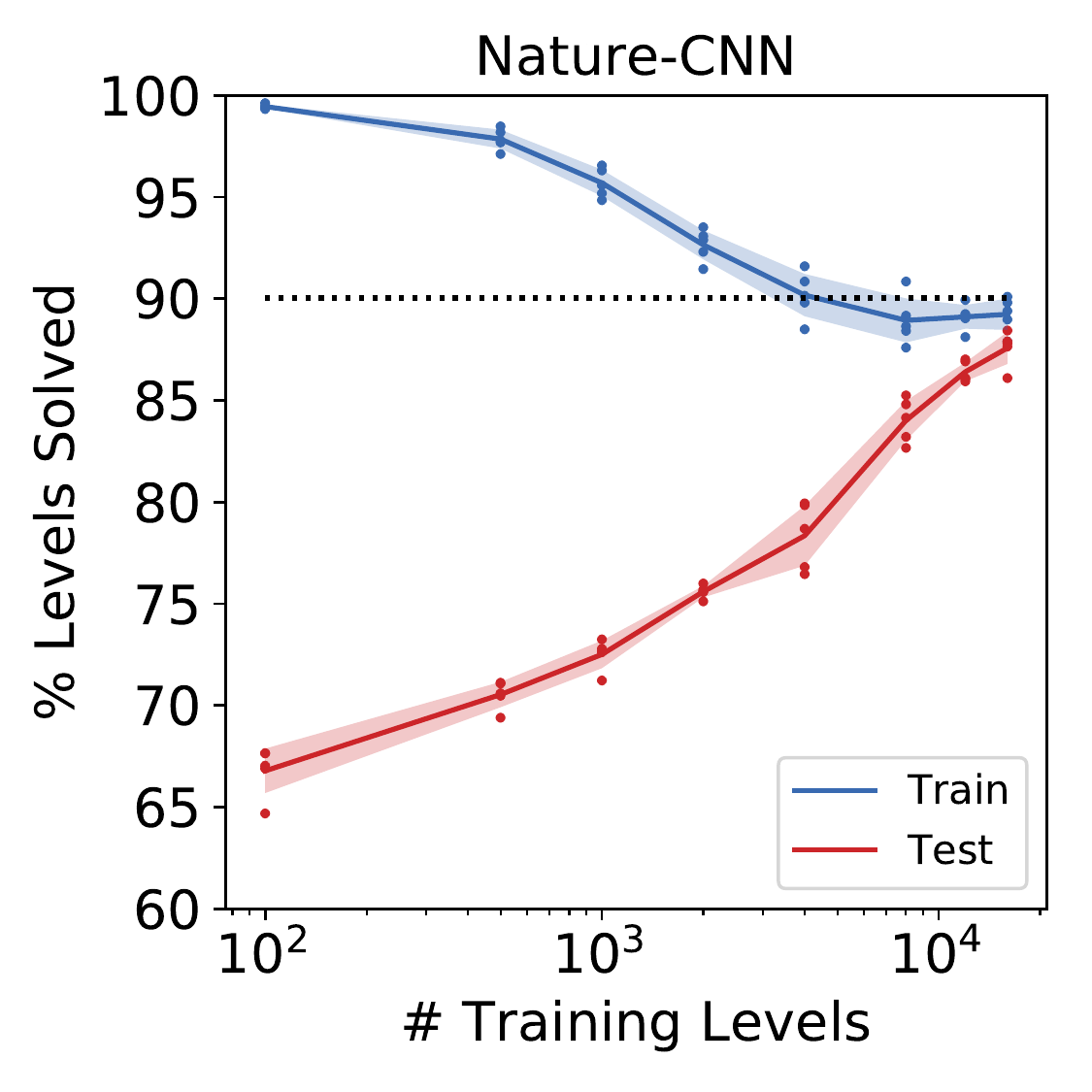}
\caption{Final train and test performance of Nature-CNN agents after 256M timesteps, as a function of the number of training levels.}
\label{fig:arch_nat}
\end{subfigure}
\hspace*{\fill}
\begin{subfigure}{0.475 \textwidth}
\includegraphics[width=\textwidth]{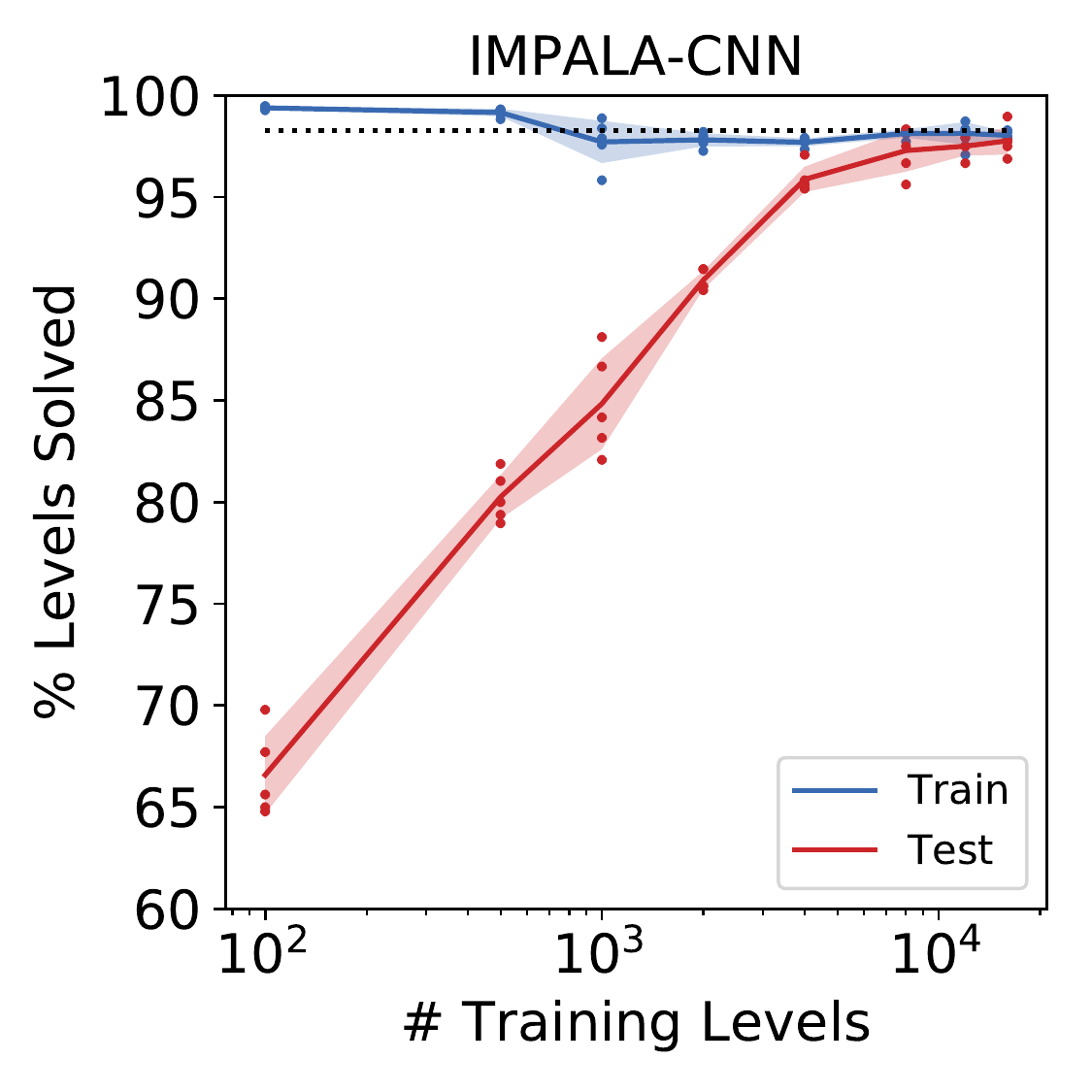}
\caption{Final train and test performance of IMPALA-CNN agents after 256M timesteps, as a function of number of training levels.} \label{fig:arch_imp}
\end{subfigure}
\caption{Dotted lines denote final mean test performance of the agents trained with an unbounded set of levels. The solid line and shaded regions represent the mean and standard deviation respectively across 5 seeds. Training sets are generated separately for each seed.}
\label{fig:all_arch}
\end{figure*}

\subsection{CoinRun Generalization Curves} \label{sec:baseline}

Using the CoinRun environment, we can measure how successfully agents generalize from a given set of training levels to an unseen set of test levels. Train and test levels are drawn from the same distribution, so the gap between train and test performance determines the extent of overfitting. As the number of available training levels grows, we expect the performance on the test set to improve, even when agents are trained for a fixed number of timesteps. At test time, we measure the zero-shot performance of each agent on the test set, applying no fine-tuning to the agent’s parameters.

We train 9 agents to play CoinRun, each on a training set with a different number of levels. During training, each new episode uniformly samples a level from the appropriate set. The first 8 agents are trained on sets ranging from of 100 to 16,000 levels. We train the final agent on an unbounded set of levels, where each level is seeded randomly. With $2^{32}$ level seeds, collisions are unlikely. Although this agent encounters approximately 2M unique levels during training, it still does not encounter any test levels until test time. We repeat this whole experiment 5 times, regenerating the training sets each time.

We first train agents with policies using the same 3-layer convolutional architecture proposed by \cite{dqn_2}, which we henceforth call Nature-CNN. Agents are trained with Proximal Policy Optimization \citep{ppo, baselines} for a total of 256M timesteps across 8 workers. We train agents for the same number of timesteps independent of the number of levels in the training set. We average gradients across all 8 workers on each mini-batch. We use $\gamma = .999$, as an optimal agent takes between 50 and 500 timesteps to solve a level, depending on level difficulty. See \Cref{appendix:hyperparameters} for a full list of hyperparameters.

Results are shown in \Cref{fig:arch_nat}. We collect each data point by averaging the final agent's performance across 10,000 episodes, where each episode samples a level from the appropriate set. We can see that substantial overfitting occurs when there are less than 4,000 training levels. Even with 16,000 training levels, overfitting is still noticeable. Agents perform best when trained on an unbounded set of levels, when a new level is encountered in every episode. See \Cref{appendix:performance} for performance details.

Now that we have generalization curves for the baseline architecture, we can evaluate the impact of various algorithmic and architectural decisions.

\begin{figure*}
\centering
\begin{subfigure}{0.475 \textwidth}
\includegraphics[width=.9\linewidth]{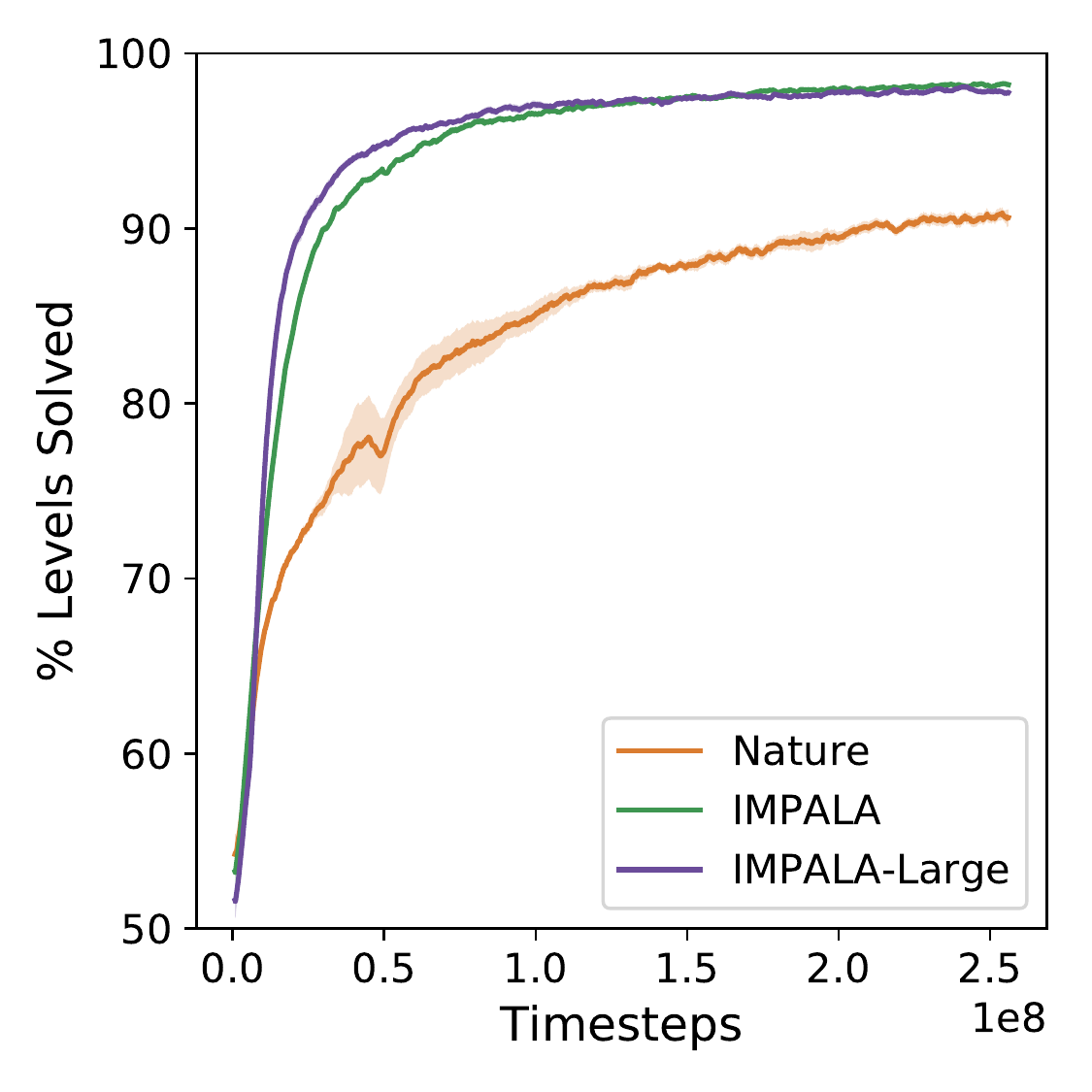}
\caption{Performance of Nature-CNN and IMPALA-CNN agents during training, on an unbounded set of training levels.}
 \label{fig:arch_gen_all}
\end{subfigure}
\hspace*{\fill}
\begin{subfigure}{0.475 \textwidth}
\includegraphics[width=.9\linewidth]{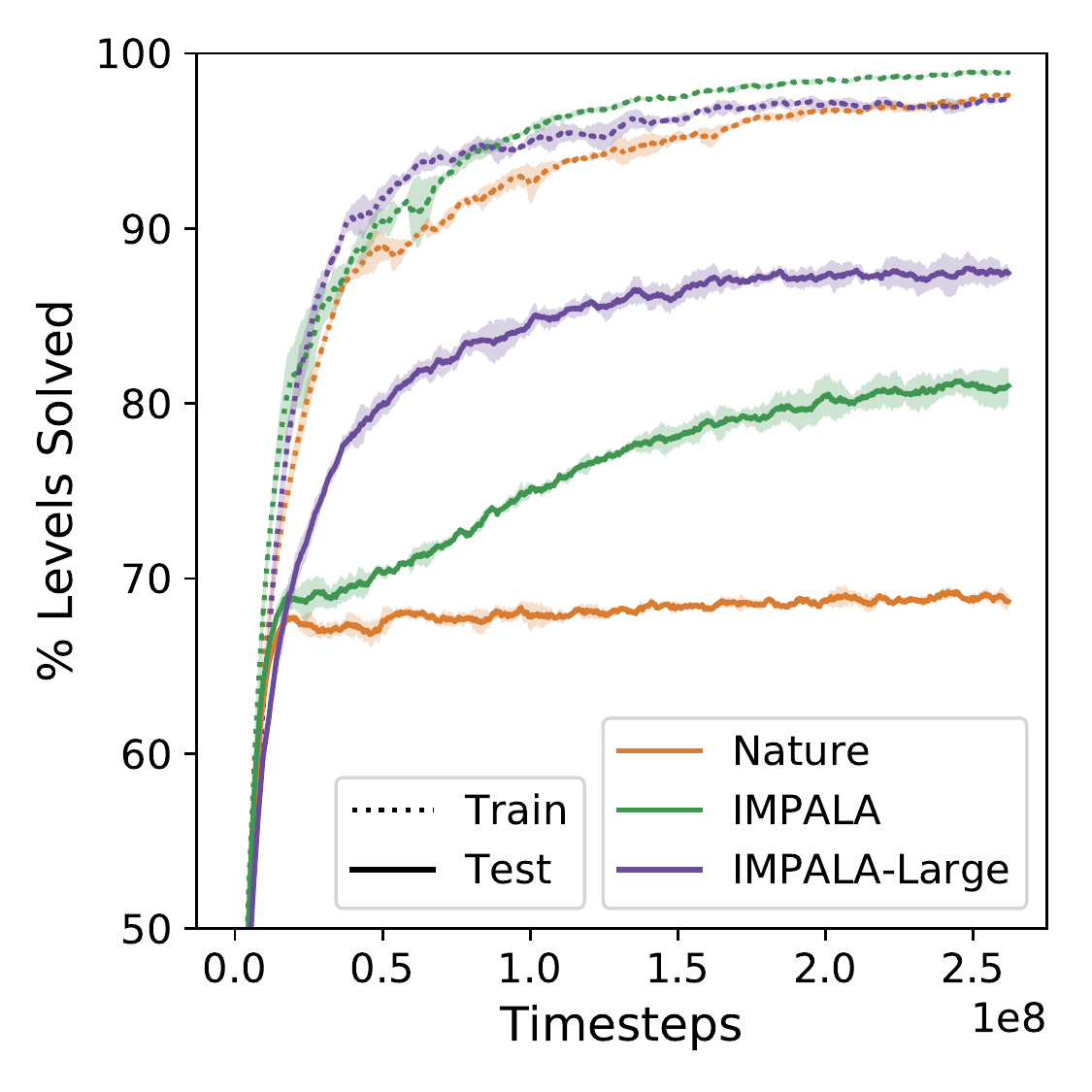}
\caption{Performance of Nature-CNN and IMPALA-CNN agents during training, on a set of 500 training levels.} \label{fig:arch_gen_500}
\end{subfigure}
\caption{The lines and shaded regions represent the mean and standard deviation respectively across 3 runs.}
\label{fig:arch_comp}
\end{figure*}

\section{Evaluating Architectures} \label{sec:eval_arch}

We choose to compare the convolutional architecture used in IMPALA \citep{impala} against our Nature-CNN baseline. With the IMPALA-CNN, we perform the same experiments described in \Cref{sec:baseline}, with results shown in \Cref{fig:arch_imp}. We can see that across all training sets, the IMPALA-CNN agents perform better at test time than Nature-CNN agents.

To evaluate generalization performance, one could train agents on the unbounded level set and directly compare learning curves. In this setting, it is impossible for an agent to overfit to any subset of levels. Since every level is new, the agent is evaluated on its ability to continually generalize. For this reason, performance with an unbounded training set can serve as a reasonable proxy for the more explicit train-to-test generalization performance. \Cref{fig:arch_gen_all} shows a comparison between training curves for IMPALA-CNN and Nature-CNN, with an unbounded set of training levels. As we can see, the IMPALA-CNN architecture is substantially more sample efficient.

However, it is important to note that learning faster with an unbounded training set will not always correlate positively with better generalization performance. In particular, well chosen hyperparameters might lead to improved training speed, but they are less likely to lead to improved generalization. We believe that directly evaluating generalization, by training on a fixed set of levels, produces the most useful metric. \Cref{fig:arch_gen_500} shows the performance of different architectures when training on a fixed set of 500 levels. The same training set is used across seeds.

In both settings, it is clear that the IMPALA-CNN architecture is better at generalizing across levels of CoinRun. Given the success of the IMPALA-CNN, we experimented with several larger architectures, finding a deeper and wider variant of the IMPALA architecture (IMPALA-Large) that performs even better. This architecture uses 5 residual blocks instead of 3, with twice as many channels at each layer. Results with this architecture are shown in \Cref{fig:arch_comp}.

It is likely that further architectural tuning could yield even greater generalization performance. As is common in supervised learning, we expect much larger networks to have a higher capacity for generalization. In our experiments, however, we noticed diminishing returns increasing the network size beyond IMPALA-Large, particularly as wall clock training time can dramatically increase. In any case, we leave further architectural investigation to future work.

\begin{figure*}
\centering
\begin{subfigure}{0.31 \textwidth}
\includegraphics[width=\textwidth]{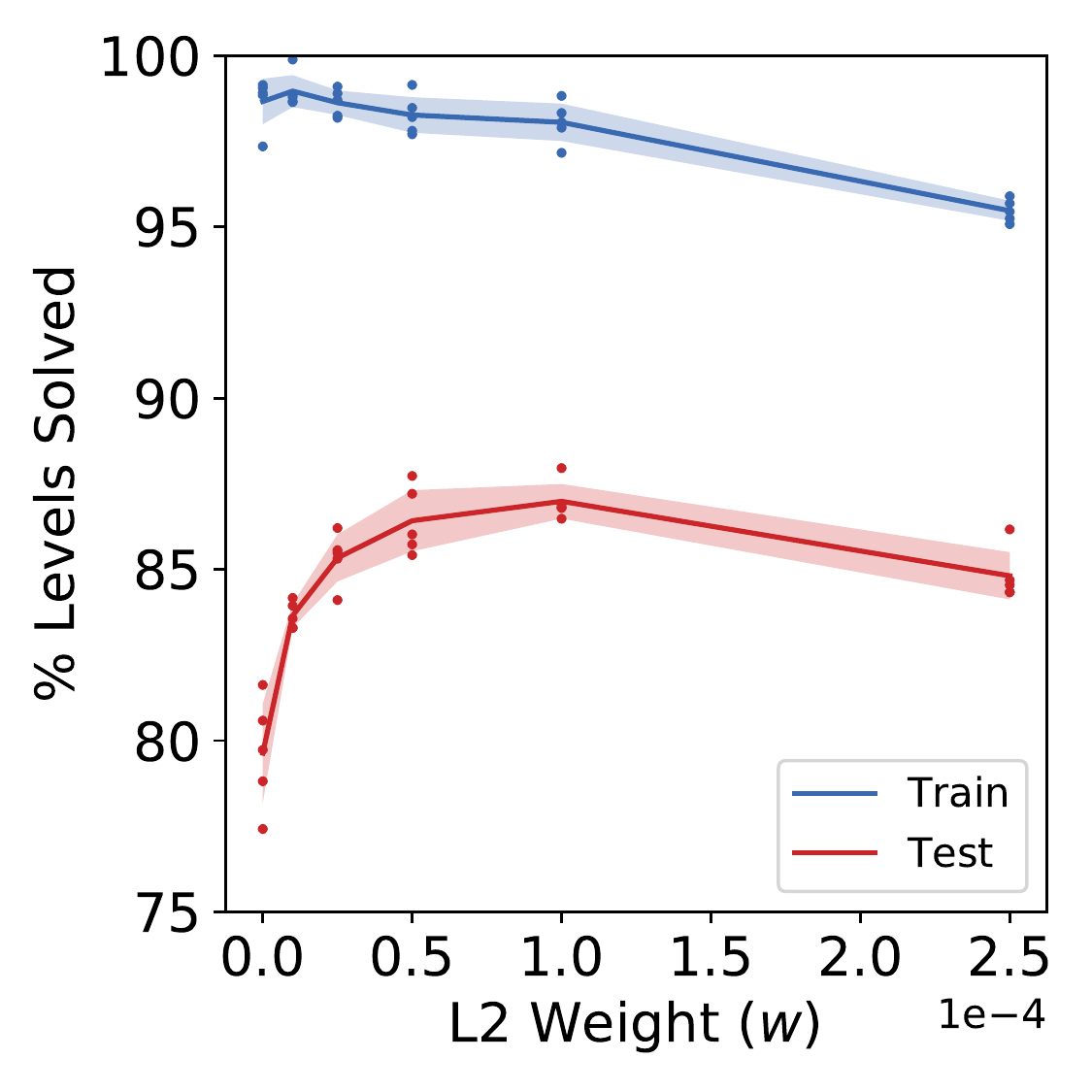}
\caption{Final train and test performance after 256M timesteps as a function of the L2 weight penalty. Mean and standard deviation is shown across 5 runs.} \label{fig:l2_reg}
\end{subfigure}
\hspace*{\fill}
\begin{subfigure}{0.31 \textwidth}
\includegraphics[width=\textwidth]{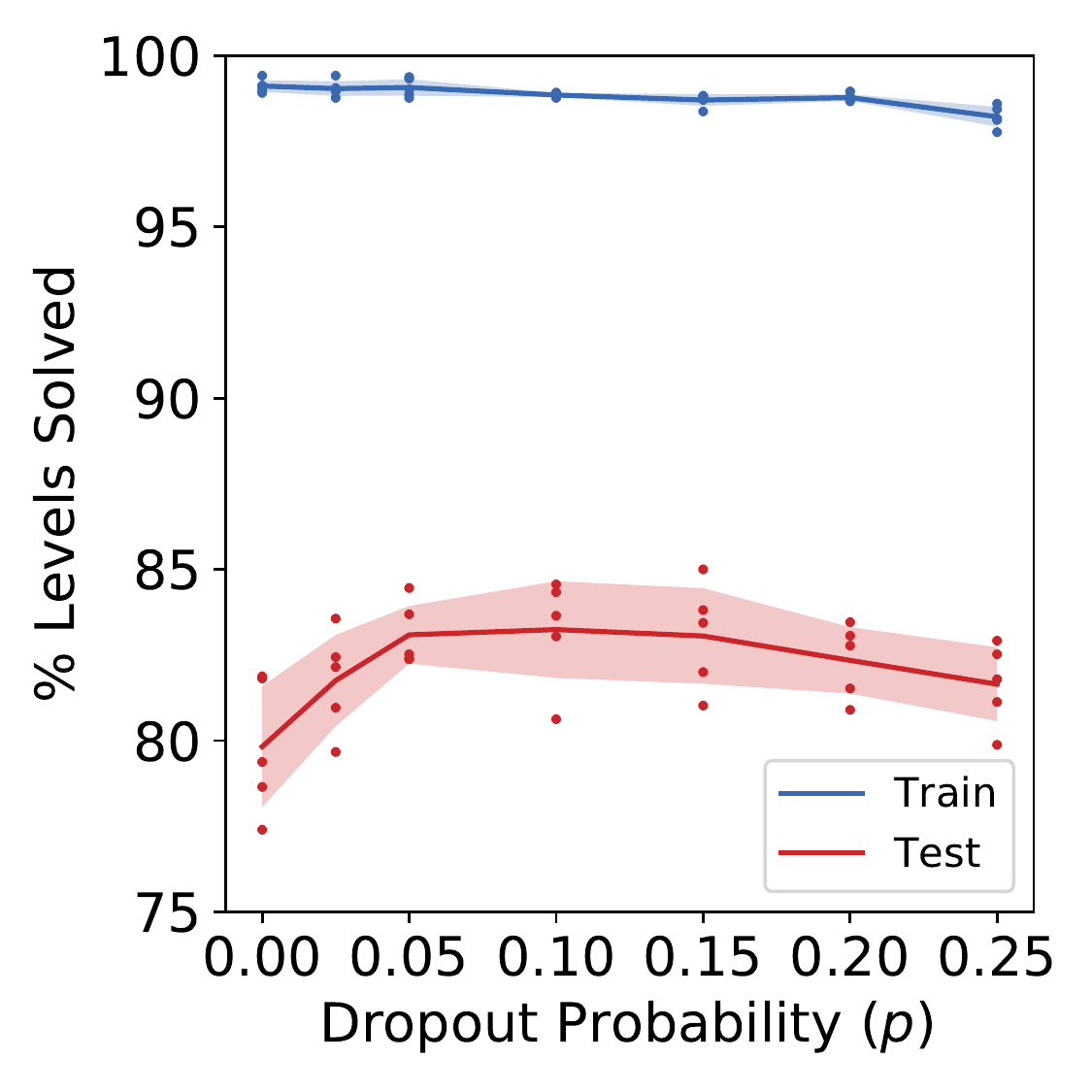}
\caption{Final train and test performance after 512M timesteps as a function of the dropout probability. Mean and standard deviation is shown across 5 runs.} \label{fig:dropout}
\end{subfigure}
\hspace*{\fill}
\begin{subfigure}{0.31 \textwidth}
\includegraphics[width=\textwidth]{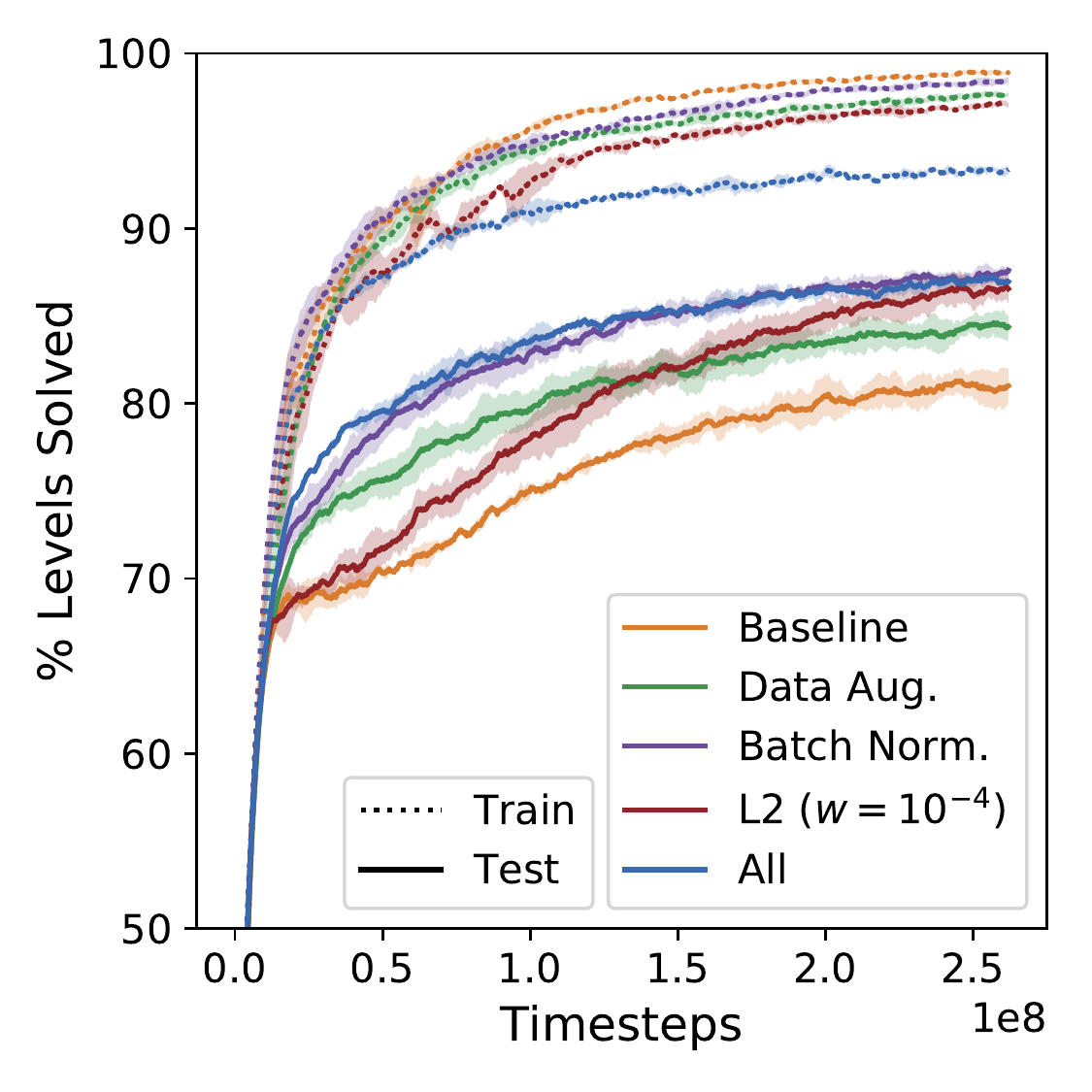}
\caption{The effect of using data augmentation, batch normalization and L2 regularization when training on 500 levels. Mean and standard deviation is shown across 3 runs.} \label{fig:misc_gen}
\end{subfigure}
\caption{The impact of different forms of regularization.}
\label{fig:all_reg}
\end{figure*}

\section{Evaluating Regularization} \label{sec:eval_all_reg}

Regularization has long played an significant role in supervised learning, where generalization is a more immediate concern. Datasets always include separate training and test sets, and there are several well established regularization techniques for reducing the generalization gap. These regularization techniques are less often employed in deep RL, presumably because they offer no perceivable benefits in the absence of a generalization gap -- that is, when the training and test sets are one and the same.

Now that we are directly measuring generalization in RL, we have reason to believe that regularization will once again prove effective. Taking inspiration from supervised learning, we choose to investigate the impact of L2 regularization, dropout, data augmentation, and batch normalization in the CoinRun environment.

Throughout this section we train agents on a fixed set of 500 CoinRun levels, following the same experimental procedure shown in \Cref{fig:arch_gen_500}. We have already seen that substantial overfitting occurs, so we expect this setting to provide a useful signal for evaluating generalization. In all subsequent experiments, figures show the mean and standard deviation across 3-5 runs. In these experiments, we use the original IMPALA-CNN architecture with 3 residual blocks, but we notice qualitatively similar results with other architectures.

\subsection{Dropout and L2 Regularization}\label{sec:eval_reg}

We first train agents with either dropout probability $p \in [0, 0.25]$ or with L2 penalty $w \in [0, \num{2.5e-4}]$. We train agents with L2 regularization for 256M timesteps, and we train agents with dropout for 512M timesteps. We do this since agents trained with dropout take longer to converge. We report both the final train and test performance. The results of these experiments are shown in \Cref{fig:all_reg}. Both L2 regularization and dropout noticeably reduce the generalization gap, though dropout has a smaller impact. Empirically, the most effective dropout probability is $p=0.1$ and the most effective L2 weight is $w=10^{-4}$.

\subsection{Data Augmentation} \label{sec:data_aug}

Data augmentation is often effective at reducing overfitting on supervised learning benchmarks. There have been a wide variety of augmentation transformations proposed for images, including translations, rotations, and adjustments to brightness, contrast, or sharpness. \cite{autoaugment} search over a diverse space of augmentations and train a policy to output effective data augmentations for a target dataset, finding that different datasets often benefit from different sets of augmentations.

We take a simple approach in our experiments, using a slightly modified form of Cutout \citep{cutout}. For each observation, multiple rectangular regions of varying sizes are masked, and these masked regions are assigned a random color. See \Cref{appendix:data_aug_screenshots} for screenshots. This method closely resembles domain randomization \citep{domain_rand}, used in robotics to transfer from simulations to the real world. \Cref{fig:misc_gen} shows the boost this data augmentation scheme provides in CoinRun. We expect that other methods of data augmentation would prove similarly effective and that the effectiveness of any given augmentation will vary across environments.

\begin{figure*}
\centering
\begin{subfigure}{0.31 \textwidth}
\includegraphics[width=\textwidth]{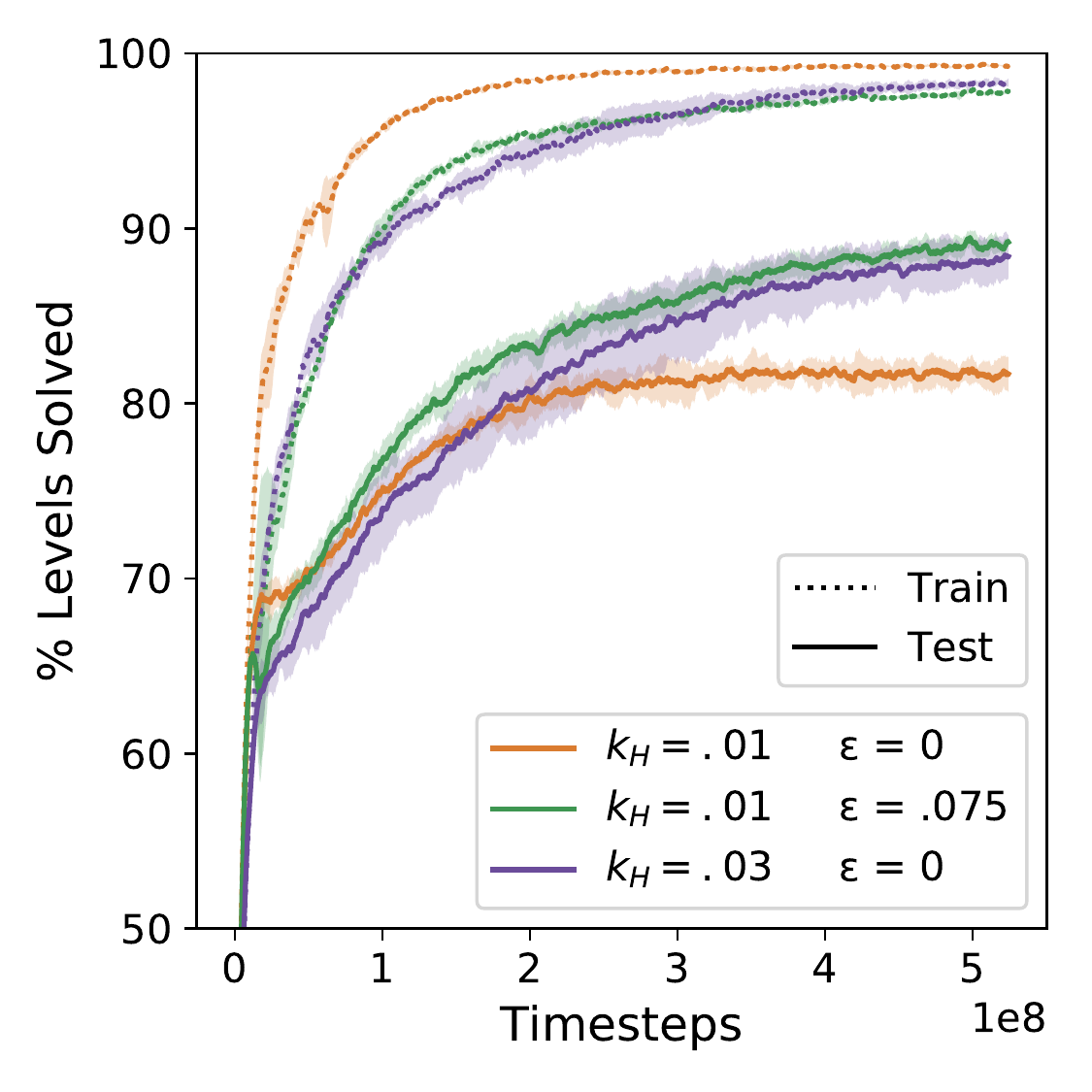}
\caption{Comparison of $\epsilon$-greedy and high entropy bonus agents to baseline during training.}
\label{fig:stoch_comp}
\end{subfigure}
\hspace*{\fill}
\begin{subfigure}{0.31 \textwidth}
\includegraphics[width=\textwidth]{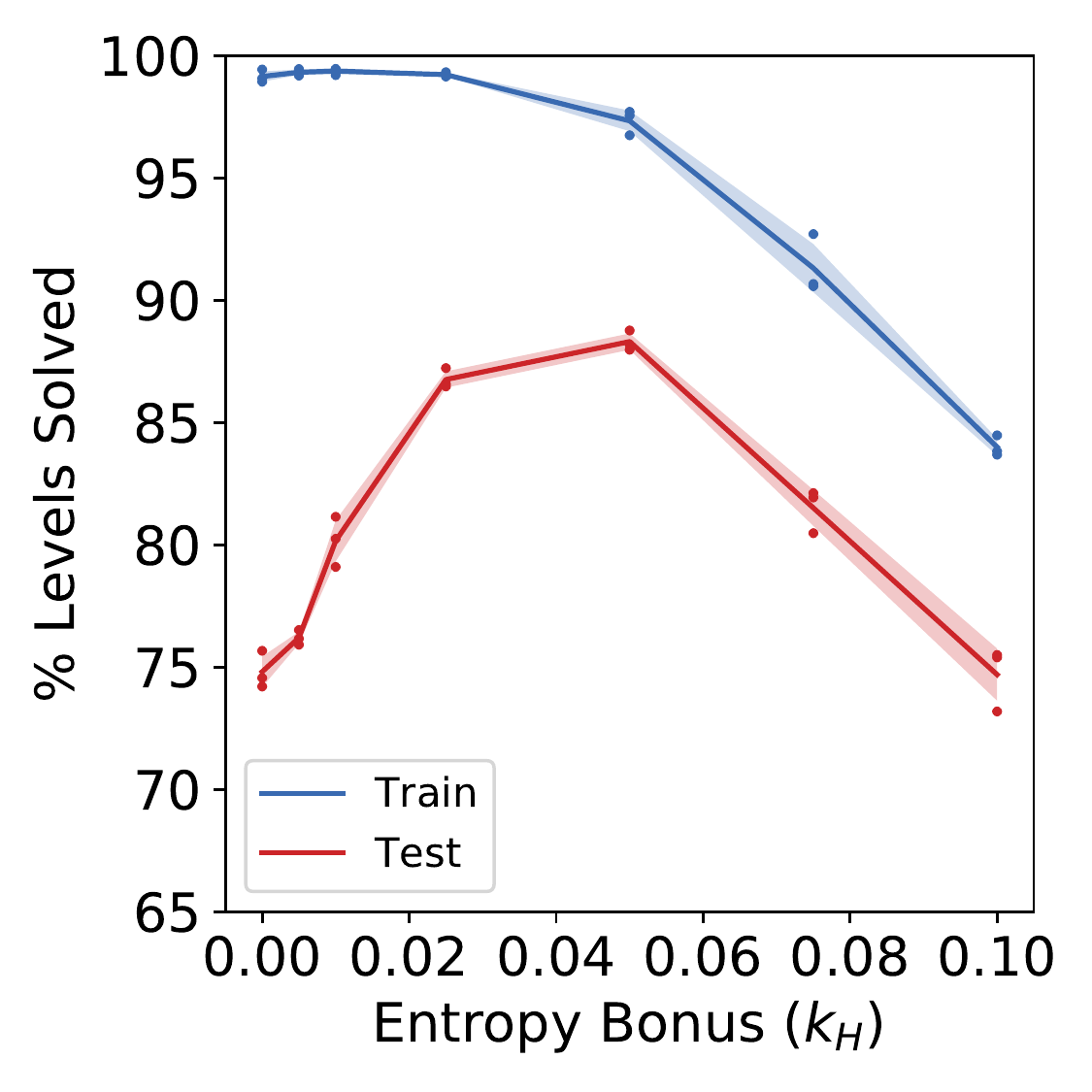}
\caption{Final train and test performance for agents trained with different entropy bonuses.} \label{fig:entropy_bonus}
\end{subfigure}
\hspace*{\fill}
\begin{subfigure}{0.31 \textwidth}
\includegraphics[width=\textwidth]{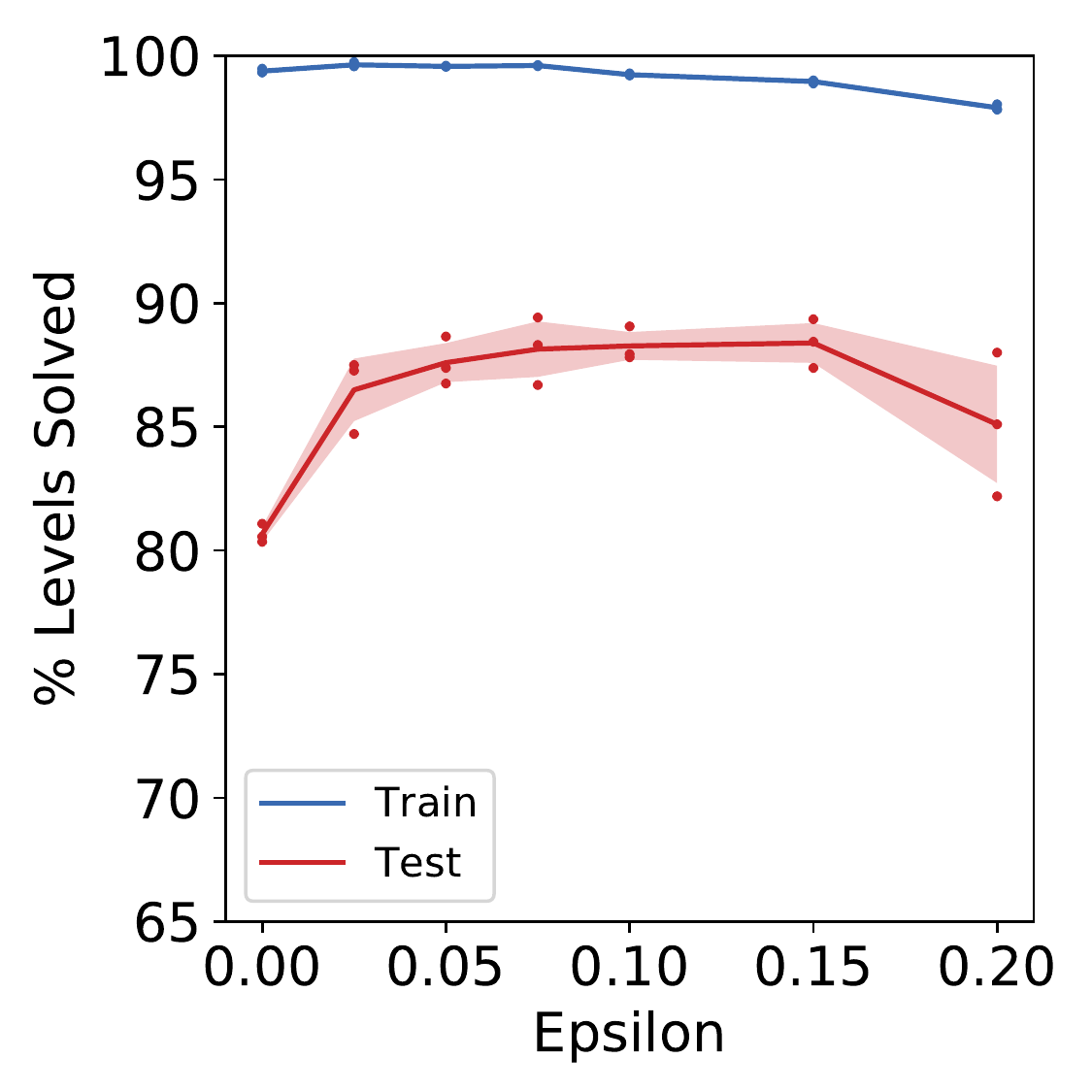}
\caption{Final train and test performance for $\epsilon$-greedy agents trained with different values of $\epsilon$.} \label{fig:random_action_b}
\end{subfigure}
\caption{The impact of introducing stochasticity into the environment, via epsilon-greedy action selection and an entropy bonus. Training occurs over 512M timesteps. Mean and standard deviation is shown across 3 runs.}
\label{fig:stoch_gen}
\end{figure*}

\subsection{Batch Normalization} \label{sec:batch_norm}

Batch normalization \citep{batchnorm} is known to have a substantial regularizing effect in supervised learning \citep{bn_reg}. We investigate the impact of batch normalization on generalization, by augmenting the IMPALA-CNN architecture with batch normalization after every convolutional layer. Training workers normalize with the statistics of the current batch, and test workers normalize with a moving average of these statistics. We show the comparison to baseline generalization in \Cref{fig:misc_gen}. As we can see, batch normalization offers a significant performance boost.

\subsection{Stochasticity} \label{sec:env_stoch}

We now evaluate the impact of stochasticity on generalization in CoinRun. We consider two methods, one varying the environment's stochasticity and one varying the policy's stochasticity. First, we inject environmental stochasticity by following $\epsilon$-greedy action selection: with probability $\epsilon$ at each timestep, we override the agent's preferred action with a random action. In previous work, $\epsilon$-greedy action selection has been used both as a means to encourage exploration and as a theoretical safeguard against overfitting \citep{ale, dqn_1}. Second, we control policy stochasticity by changing the entropy bonus in PPO. Note that our baseline agent already uses an entropy bonus of $k_H = .01$.

We increase training time to 512M timesteps as training now proceeds more slowly. Results are shown in \Cref{fig:stoch_gen}. It is clear that an increase in either the environment's or the policy's stochasticity can improve generalization. Furthermore, each method in isolation offers a similar generalization boost. It is notable that training with increased stochasticity improves generalization to a greater extent than any of the previously mentioned regularization methods. In general, we expect the impact of these stochastic methods to vary substantially between environments; we would expect less of a boost in environments whose dynamics are already highly stochastic.

\subsection{Combining Regularization Methods} \label{sec:combinations}

We briefly investigate the effects of combining several of the aforementioned techniques. Results are shown in \Cref{fig:misc_gen}. We find that combining data augmentation, batch normalization, and L2 regularization yields slightly better test time performance than using any one of them individually. However, the small magnitude of the effect suggests that these regularization methods are perhaps addressing similar underlying causes of poor generalization. Furthermore, for unknown reasons, we had little success combining $\epsilon$-greedy action selection and high entropy bonuses with other forms of regularization.

\section{Additional Environments}

The preceding sections have revealed the high degree overfitting present in one particular environment. We corroborate these results by quantifying overfitting on two additional environments: a CoinRun variant called CoinRun-Platforms and a simple maze navigation environment called RandomMazes.

We apply the same experimental procedure described in \Cref{sec:baseline} to both CoinRun-Platforms and RandomMazes, to determine the extent of overfitting. We use the original IMPALA-CNN architecture followed by an LSTM \citep{lstm}, as memory is necessary for the agent to explore optimally. These experiments further reveal how susceptible our algorithms are to overfitting.

\begin{figure}[!h]
\centering
\includegraphics[height=3.6cm]{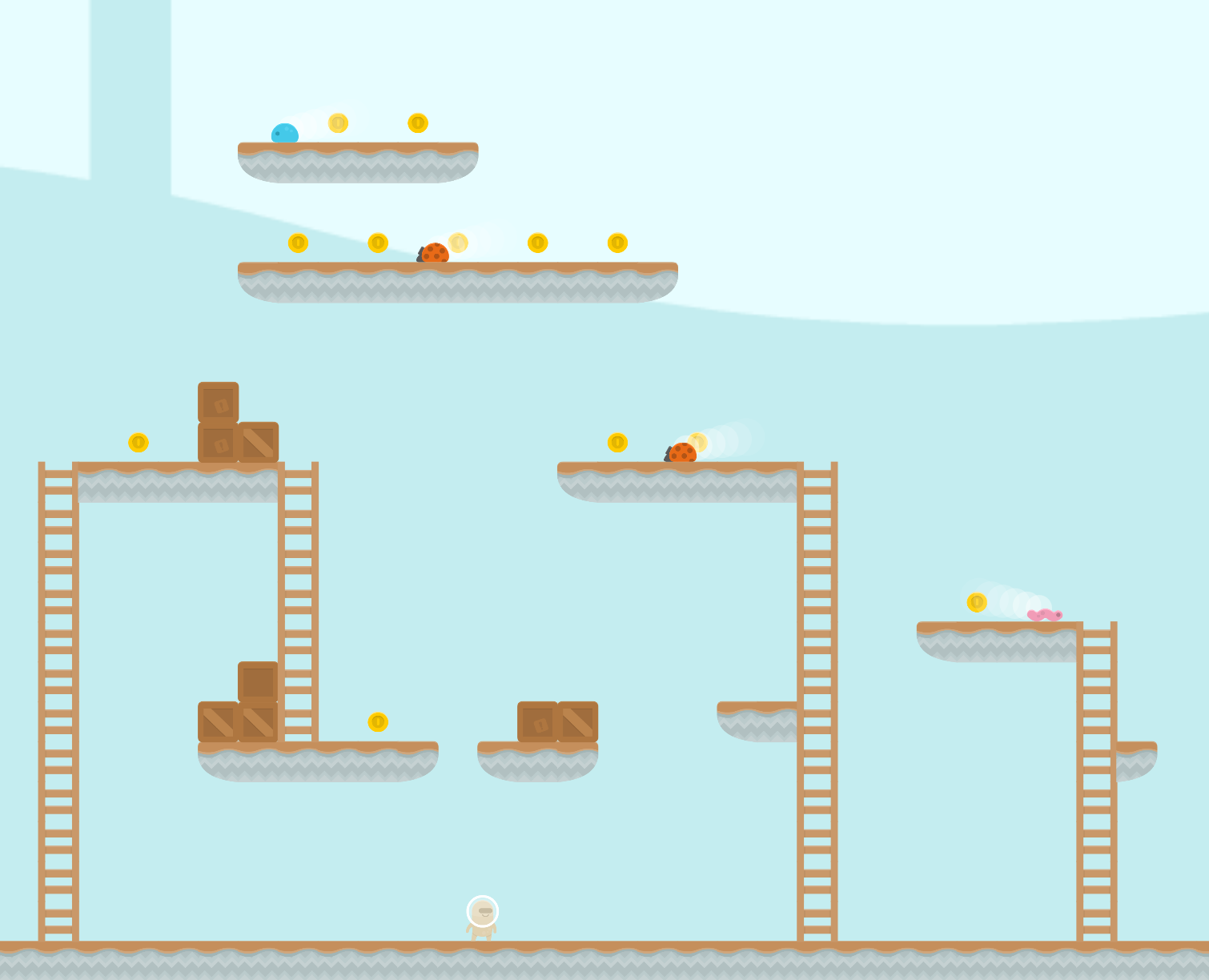}
\hspace*{\fill}
\includegraphics[height=3.6cm]{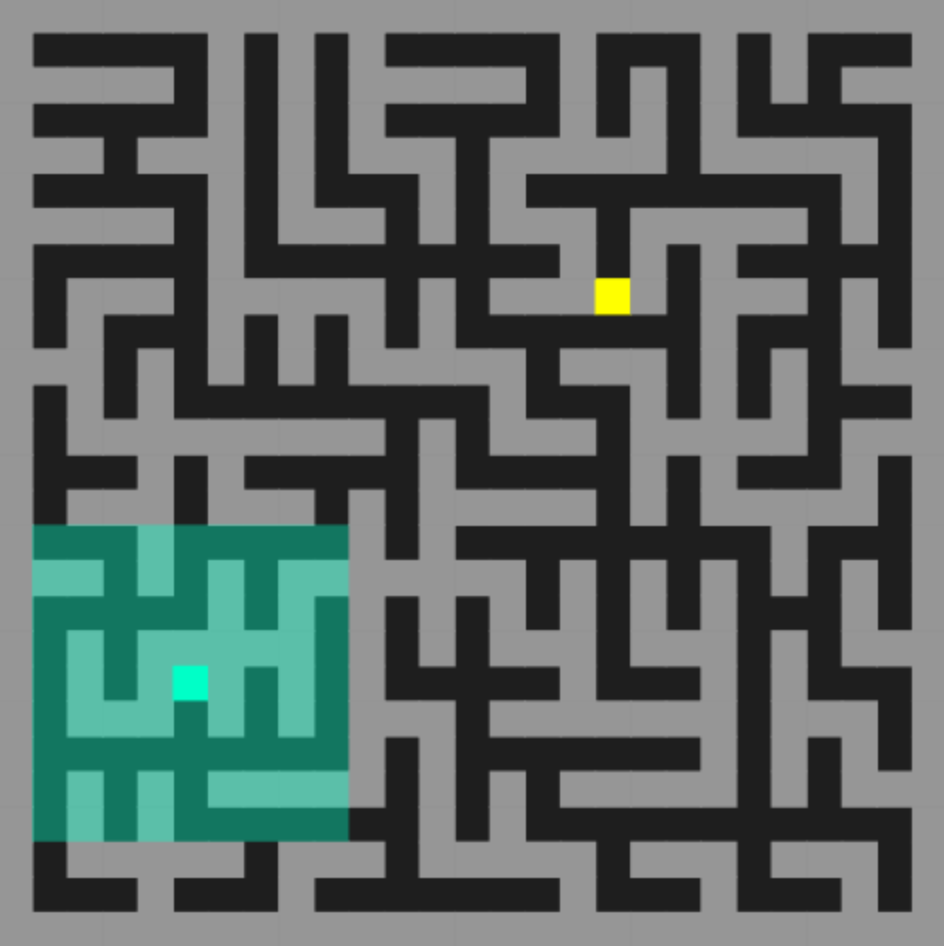}
\caption{Levels from CoinRun-Platforms (left) and RandomMazes (right). In RandomMazes, the agent's observation space is shaded in green.}
\end{figure}

\begin{figure*}
\centering
\begin{minipage}{.475\textwidth}
  \centering
  \includegraphics[width=.75\linewidth]{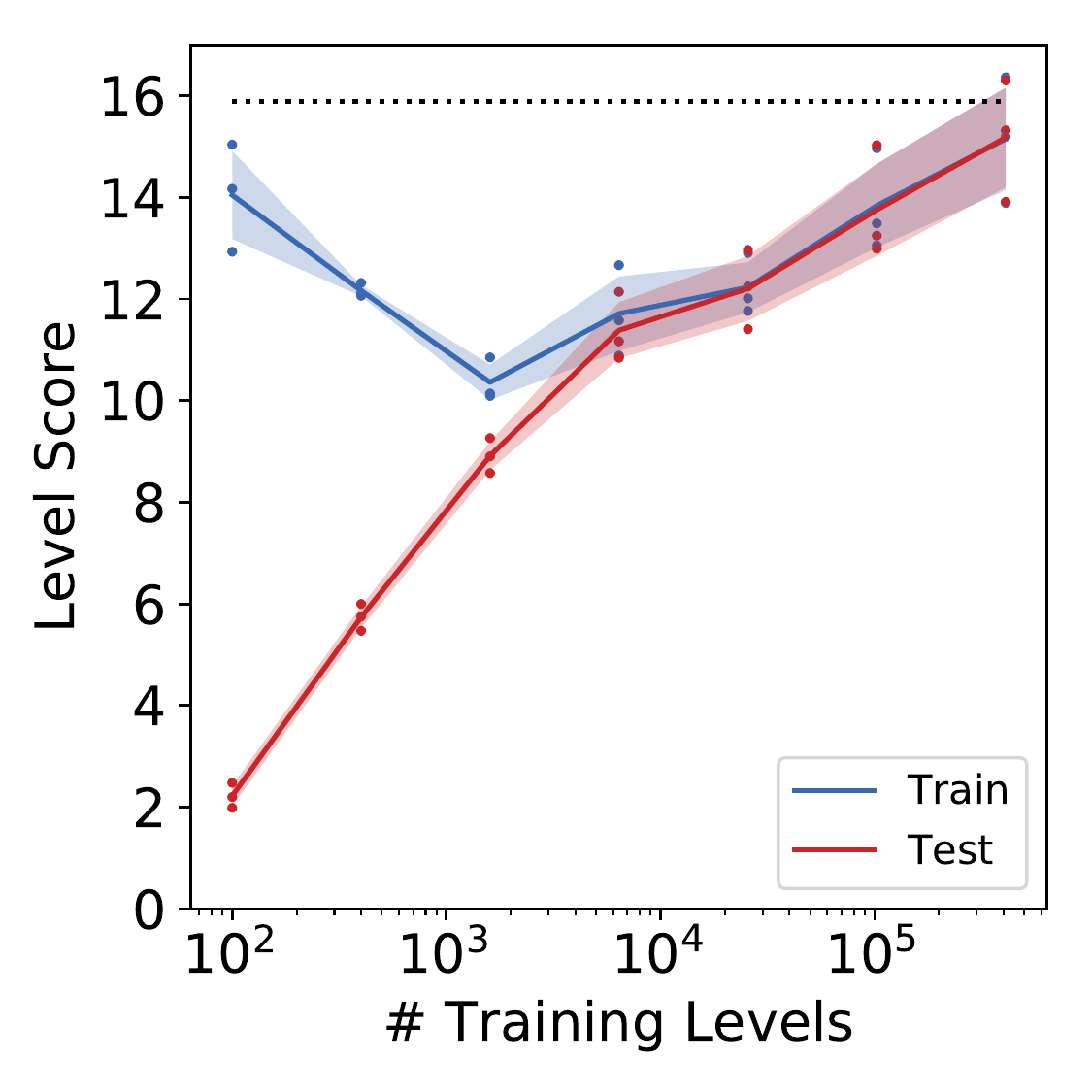}
  \captionof{figure}{Final train and test performance in CoinRun-Platforms after 2B timesteps, as a function of the number of training levels.}
  \label{fig:coinrun_plat_gen}
\end{minipage}%
\hspace*{\fill}
\begin{minipage}{.475\textwidth}
  \centering
  \includegraphics[width=.75\linewidth]{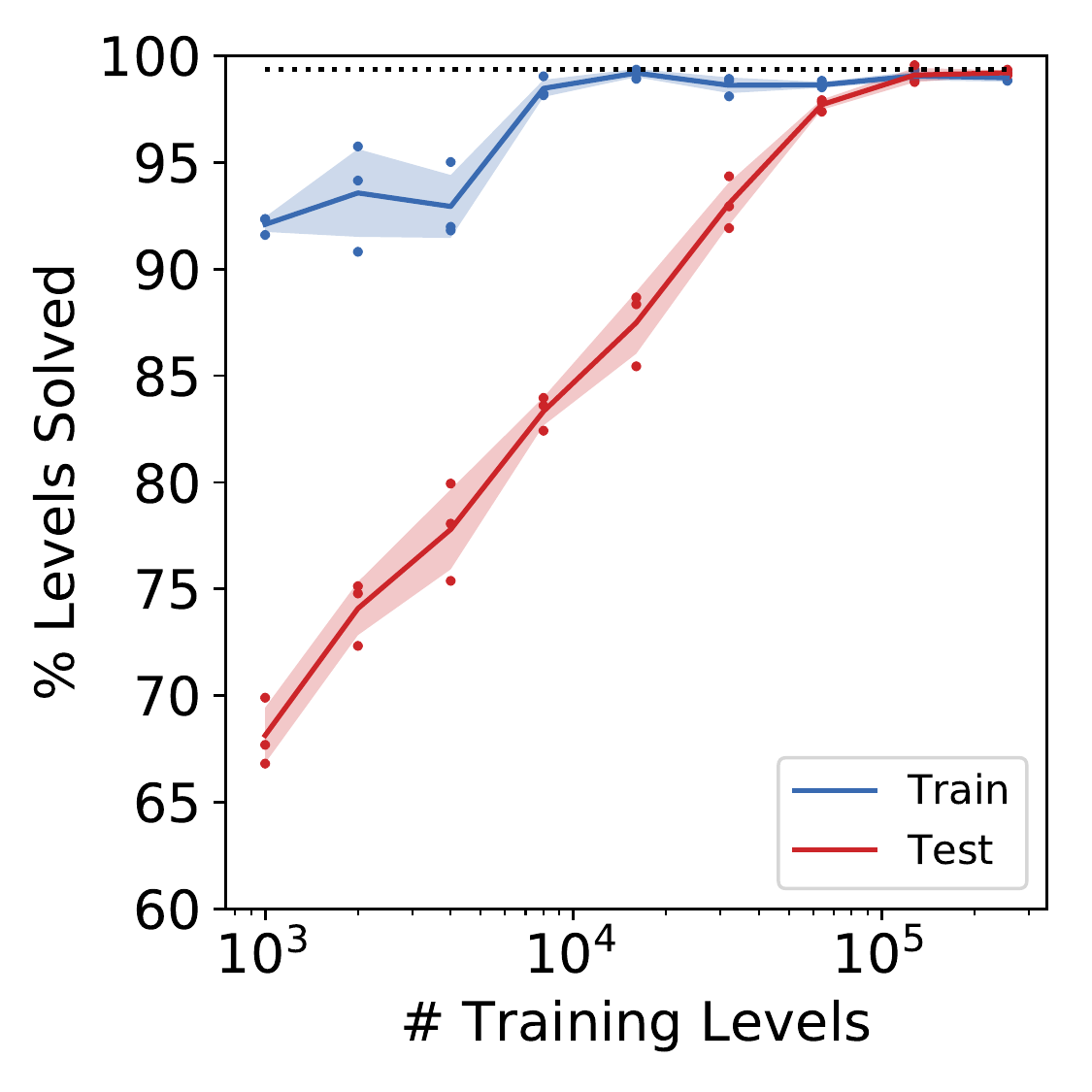}
  \captionof{figure}{Final train and test performance in RandomMazes after 256M timesteps, as a function of the number of training levels.}
  \label{fig:random_mazes}
\end{minipage}
\end{figure*}

\subsection{CoinRun-Platforms}

In CoinRun-Platforms, there are several coins that the agent attempts to collect within the 1000 step time-limit. Coins are randomly scattered across platforms in the level. Levels are a larger than in CoinRun, so the agent must actively explore, sometimes retracing its steps. Collecting any coin gives a reward of 1, and collecting all coins in a level gives an additional reward of 9. Each level contains several moving monsters that the agent must avoid. The episode ends only when all coins are collected, when time runs out, or when the agent dies. See \Cref{appendix:env_screenshots} for environment screenshots.

As CoinRun-Platforms is a much harder game, we train each agent for a total of 2B timesteps. \Cref{fig:coinrun_plat_gen} shows that overfitting occurs up to around 4000 training levels. Beyond the extent of overfitting, it is also surprising that agents' training performance increases as a function of the number of training levels, past a certain threshold. This is notably different from supervised learning, where training performance generally decreases as the training set becomes larger. We attribute this trend to the implicit curriculum in the distribution of generated levels. With additional training data, agents are more likely to learn skills that generalize even across training levels, thereby boosting the overall training performance.

\subsection{RandomMazes}

In RandomMazes, each level consists of a randomly generated square maze with dimension uniformly sampled from 3 to 25. Mazes are generated using Kruskal's algorithm \citep{kruskal}. The environment is partially observed, with the agent observing the $9\times9$ patch of cells directly surrounding its current location. At every cell is either a wall, an empty space, the goal, or the agent. The episode ends when the agent reaches the goal or when time expires after 500 timesteps. The agent's only actions are to move to an empty adjacent square. If the agent reaches the goal, a constant reward is received. \Cref{fig:random_mazes} reveals particularly strong overfitting, with a sizeable generalization gap even when training on 20,000 levels.

\subsection{Discussion}

In both CoinRun-Platforms and RandomMazes, agents must learn to leverage recurrence and memory to optimally navigate the environment. The need to memorize and recall past experience presents challenges to generalization unlike those seen in CoinRun. It is unclear how well suited LSTMs are to this task. We empirically observe that given sufficient data and training time, agents using LSTMs eventually converge to a near optimal policy. However, the relatively poor generalization performance raises the question of whether different recurrent architectures might be better suited for generalization in these environments. This investigation is left for future work.

\section{Conclusion}

Our results provide insight into the challenges underlying generalization in RL. We have observed the surprising extent to which agents can overfit to a fixed training set. Using the procedurally generated CoinRun environment, we can precisely quantify such overfitting. With this metric, we can better evaluate key architectural and algorithmic decisions. We believe that the lessons learned from this environment will apply in more complex settings, and we hope to use this benchmark, and others like it, to iterate towards more generalizable agents.

\changeurlcolor{black}

\bibliography{quantifying_generalization}

\changeurlcolor{blue}

\appendix

\clearpage
\section{Level Generation and Environment Details} \label{appendix:game_mech}

\subsection{CoinRun}

Each CoinRun level has a difficulty setting from 1 to 3. To generate a new level, we first uniformly sample over difficulties. Several choices throughout the level generation process are conditioned on this difficulty setting, including the number of sections in the level, the length and height of each section, and the frequency of obstacles. We find that conditioning on difficulty in this way creates a distribution of levels that forms a useful curriculum for the agent. For more efficient training, one could adjust the difficulty of sampled levels based on the agent’s current skill, as done in \cite{illuminating_gen}. However, we chose not to do so in our experiments, for the sake of simplicity.

At each timestep, the agent receives a $64\times64\times3$ RGB observation, centered on the agent. Given the game mechanics, an agent must know it's current velocity in order to act optimally. This requirement can be satisfied by using frame stacking or by using a recurrent model. Alternatively, we can include velocity information in each observation by painting two small squares in the upper left corner, corresponding to $x$ and $y$ velocity. In practice, agents can adequately learn under any of these conditions. Directly painting velocity information leads to the fastest learning, and we report results on CoinRun using this method. We noticed similar qualitative results using frame stacking or recurrent models, though with unsurprisingly diminished generalization performance.

\subsection{CoinRun-Platforms}

As with CoinRun, agents receive a $64\times64\times3$ RGB observation at each timestep in CoinRun-Platforms. We don't paint any velocity information into observations in experiments with CoinRun-Platforms; this information can be encoded by the LSTM used in these experiments. Unlike CoinRun, levels in CoinRun-Platforms have no explicit difficulty setting, so all levels are drawn from the same distribution. In practice, of course, some levels will still be drastically easier than others.

It is worth emphasizing that CoinRun platforms is a much more difficult game than CoinRun. We trained for 2B timesteps in \Cref{fig:coinrun_plat_gen}, but even this was not enough time for training to fully converge. We found that with unrestricted training levels, training converged after approximately 6B timesteps at a mean score of 20 per level. Although this agent occasionally makes mistakes, it appears to be reasonably near optimal performance. In particular, it demonstrates a robust ability to explore the level.

\subsection{RandomMazes}

As with the previous environments, agents receive a $64\times64\times3$ RGB observation at each timestep in RandomMazes. Given the visual simplicity of the environment, using such a larger observation space is clearly not necessary. However, we chose to do so for the sake of consistency. We did conduct experiments with smaller observation spaces, but we noticed similar levels of overfitting.

\onecolumn
\clearpage
\section{Environment Screenshots} \label{appendix:env_screenshots}

\hspace{.05\textwidth}CoinRun Difficulty 1 Levels:

\begin{figure*}[!h]
\centering
\resizebox{.9\textwidth}{!}{%
\includegraphics[height=2.92cm]{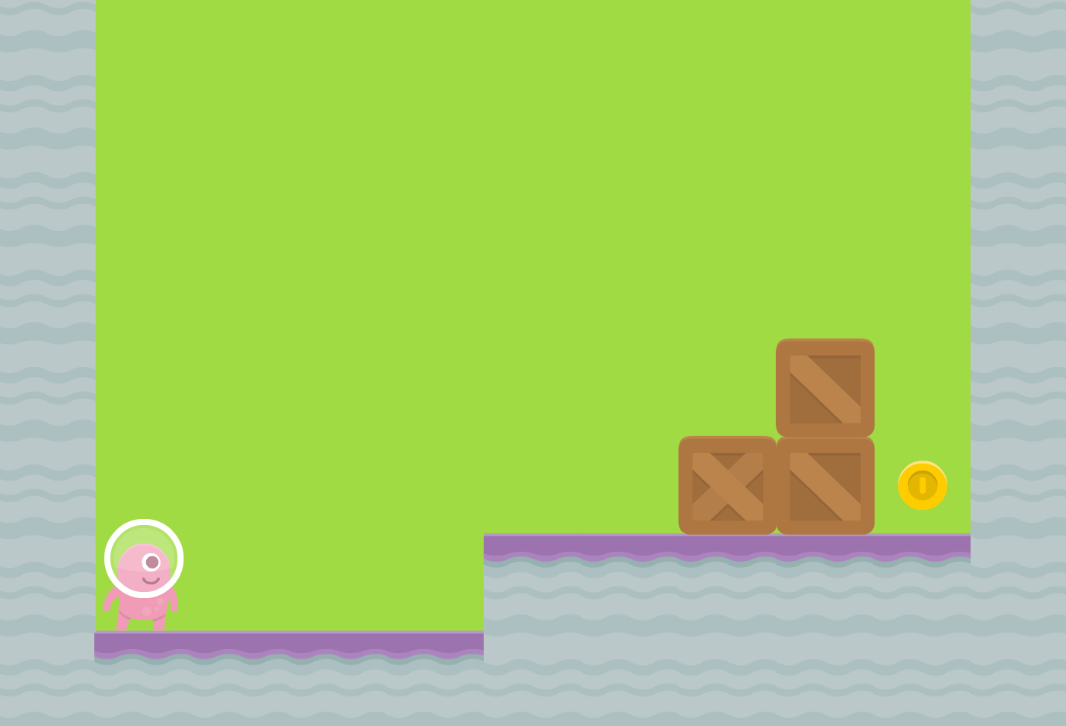}
\hspace*{\fill}
\includegraphics[height=2.92cm]{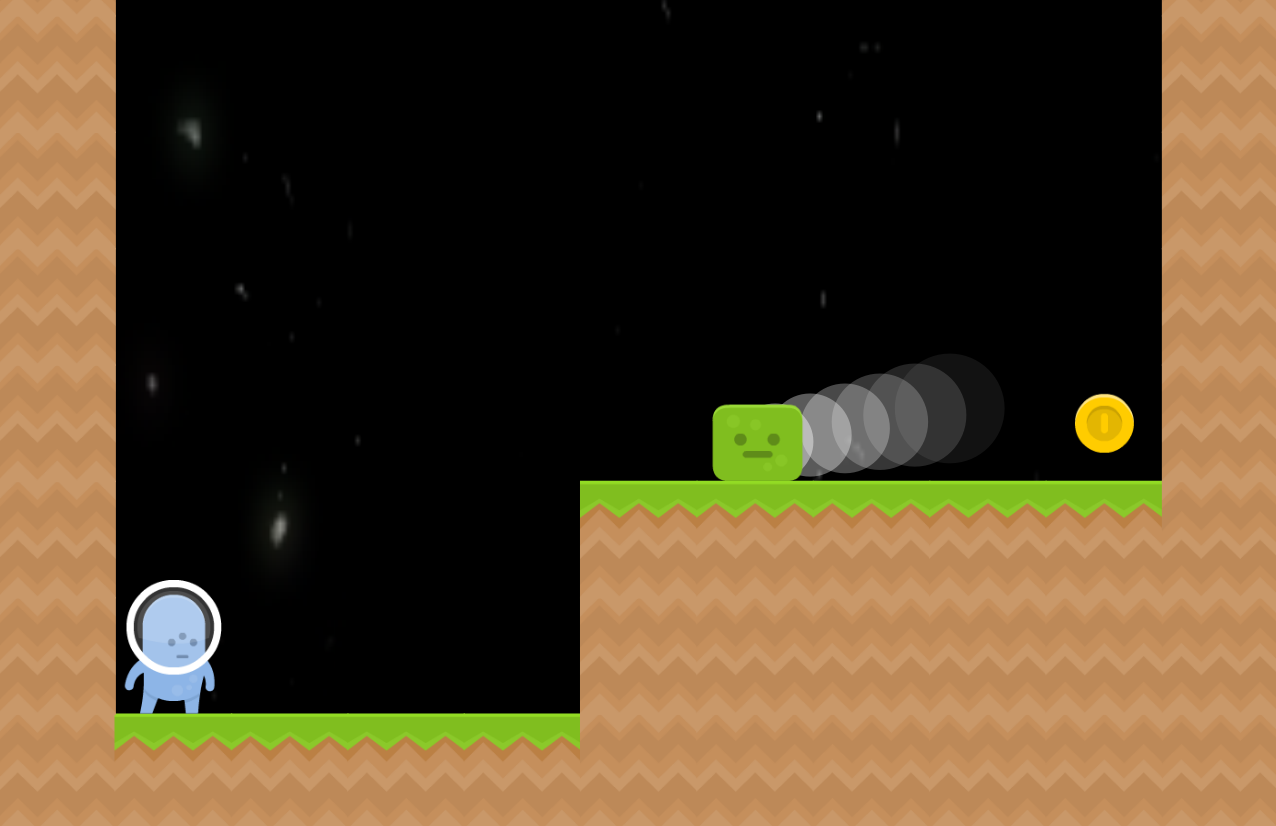}
}
\resizebox{.9\textwidth}{!}{%
\includegraphics[height=2.92cm]{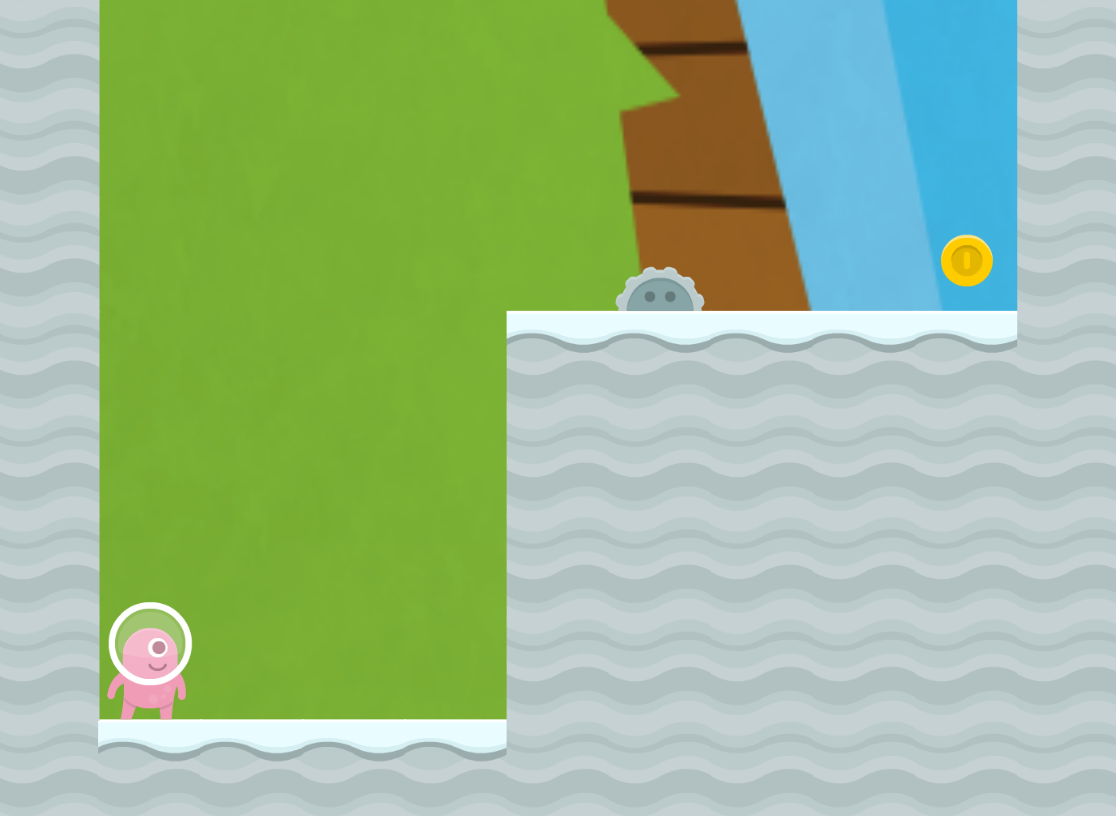}
\hspace*{\fill}
\includegraphics[height=2.92cm]{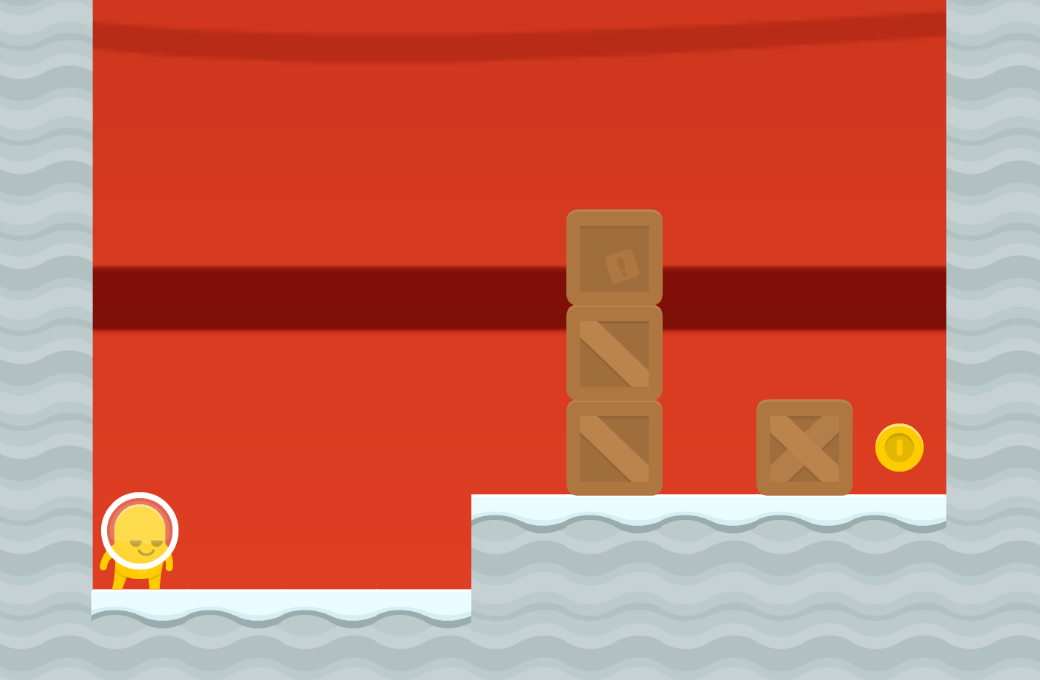}
}
\end{figure*}

\noindent
\hspace{.05\textwidth}CoinRun Difficulty 2 Levels:

\begin{figure*}[!h]
\centering
\resizebox{.9\textwidth}{!}{%
\includegraphics[height=2.7cm]{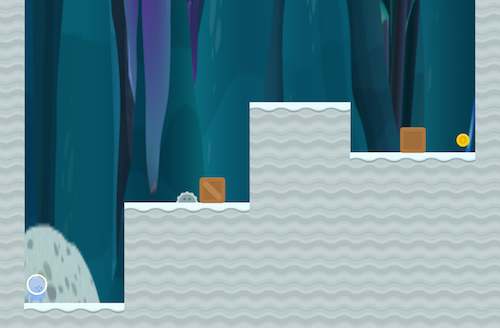}
\hspace*{\fill}
\includegraphics[height=2.7cm]{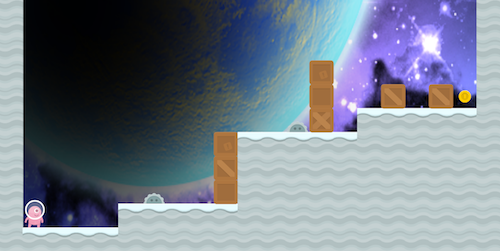}
}
\resizebox{.9\textwidth}{!}{%
\includegraphics[height=2.7cm]{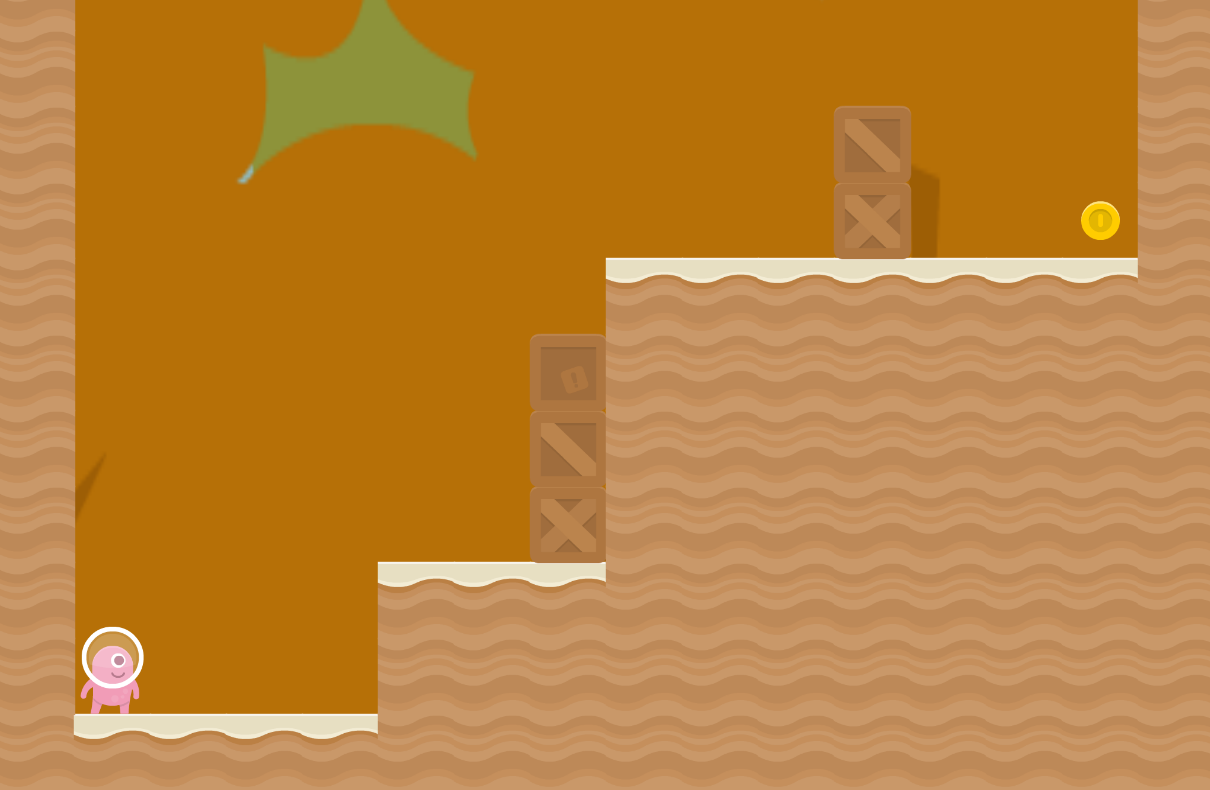}
\hspace*{\fill}
\includegraphics[height=2.7cm]{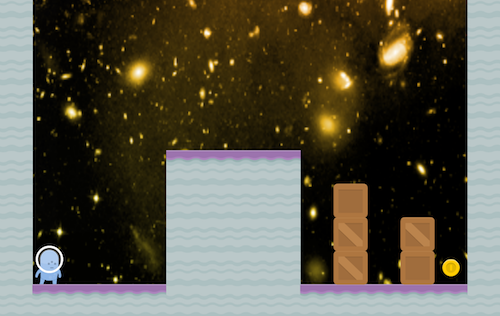}
}
\end{figure*}

\noindent
\hspace{.05\textwidth}Note: Images scaled to fit.

\newpage

\noindent
\hspace{.08\textwidth}CoinRun Difficulty 3 Levels:

\begin{figure*}[!h]
\centering
\includegraphics[width=.84\textwidth]{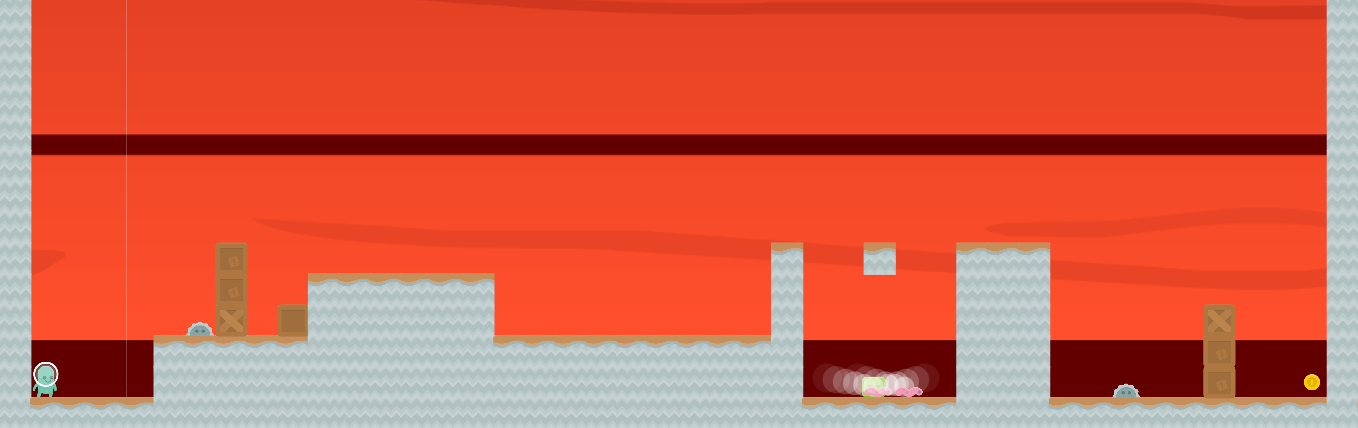}
\includegraphics[width=.84\textwidth]{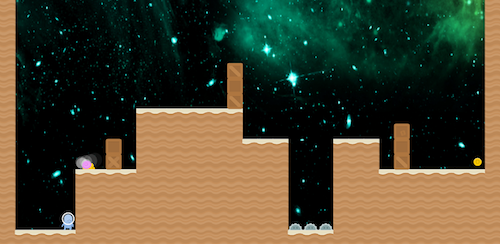}
\includegraphics[width=.84\textwidth]{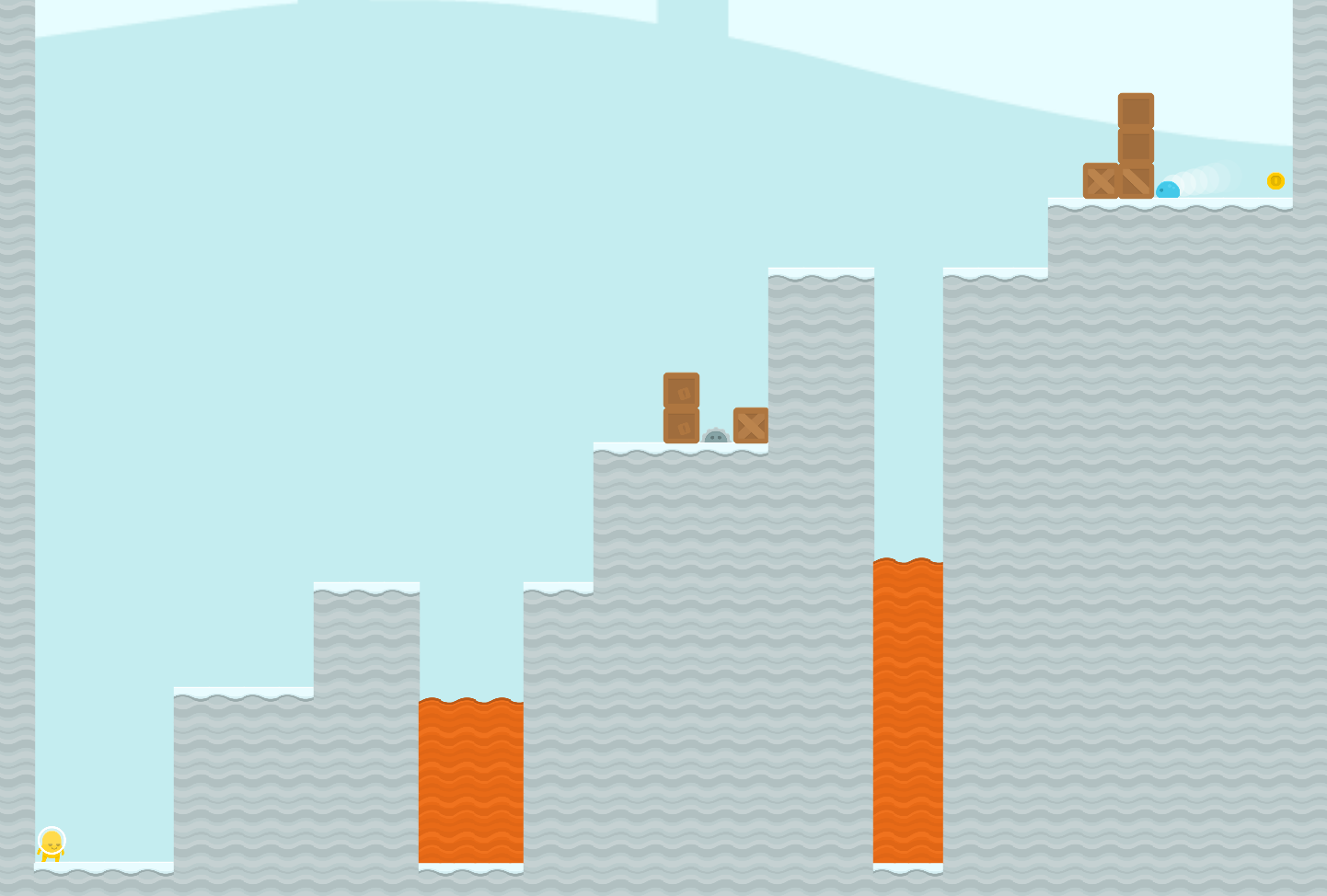}
\end{figure*}

\noindent
\hspace{.08\textwidth}Note: Images scaled to fit.

\newpage

\noindent
\hspace{.125\textwidth}CoinRun-Platforms Levels:

\begin{figure*}[!h]
\centering
\includegraphics[width=.75\textwidth]{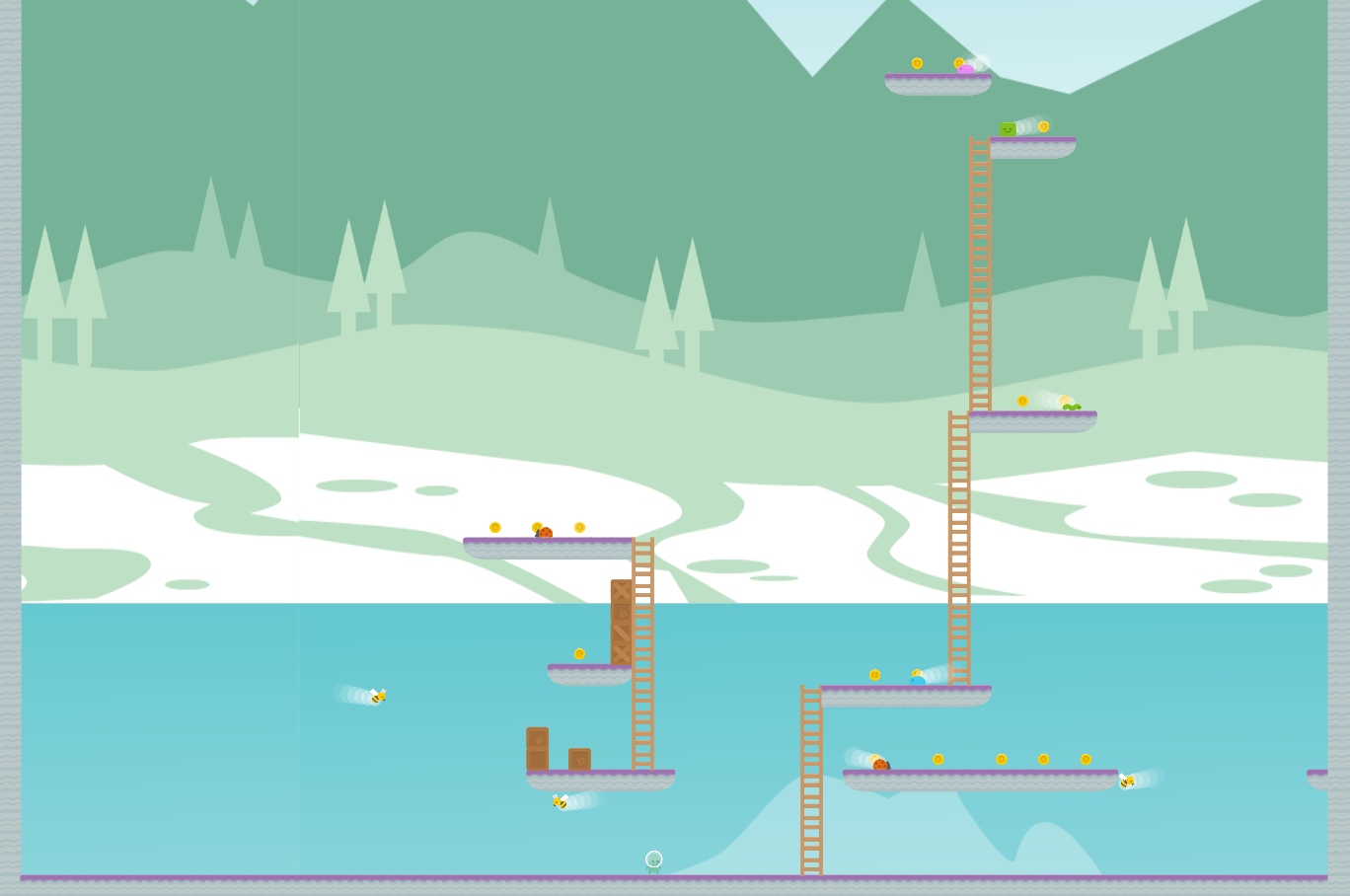}
\includegraphics[width=.75\textwidth]{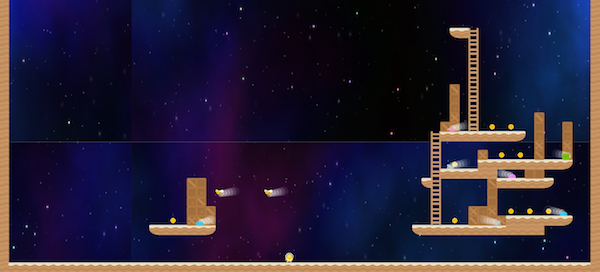}
\includegraphics[width=.75\textwidth]{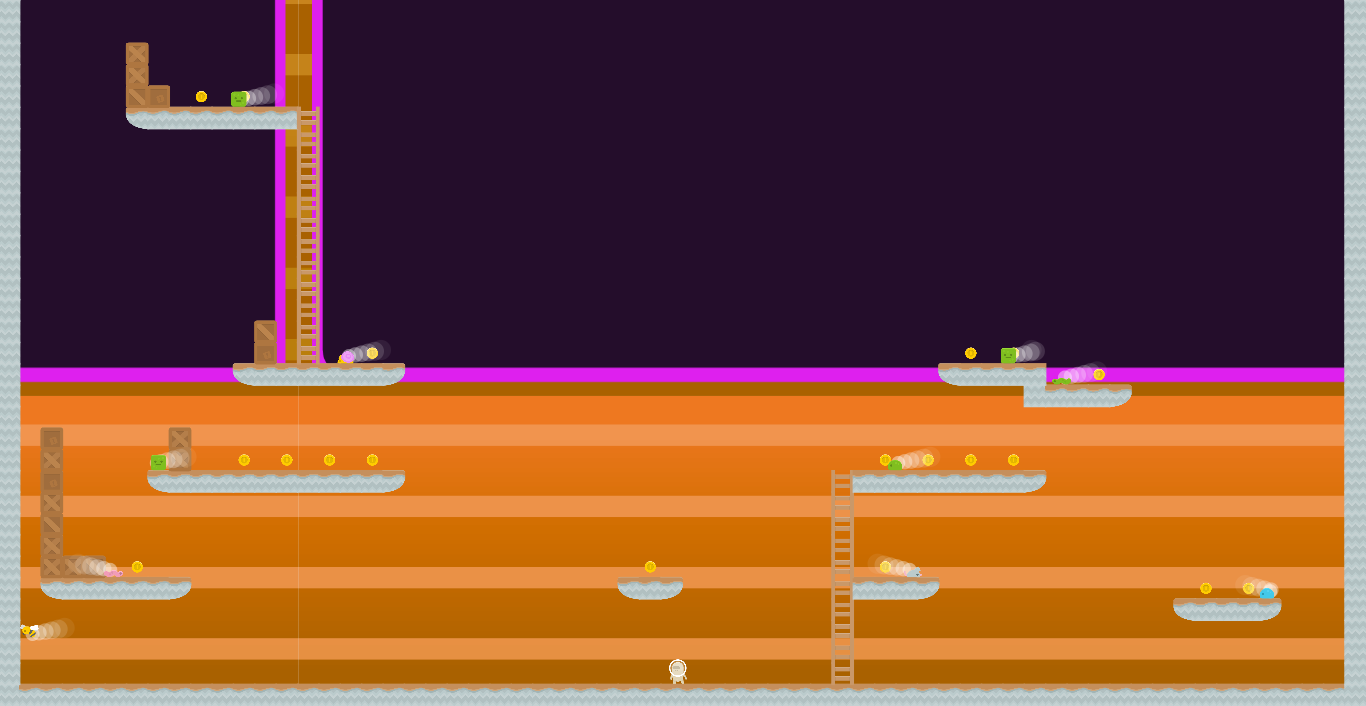}
\end{figure*}

\noindent
\hspace{.125\textwidth}Note: Images scaled to fit.

\clearpage
\section{Data Augmentation Screenshots} \label{appendix:data_aug_screenshots}

Example observations augmented with our modified version of Cutout \citep{cutout}:

\begin{figure}[!h]
\centering
\begin{minipage}{.475 \textwidth}
  \centering
  \includegraphics[width=\textwidth]{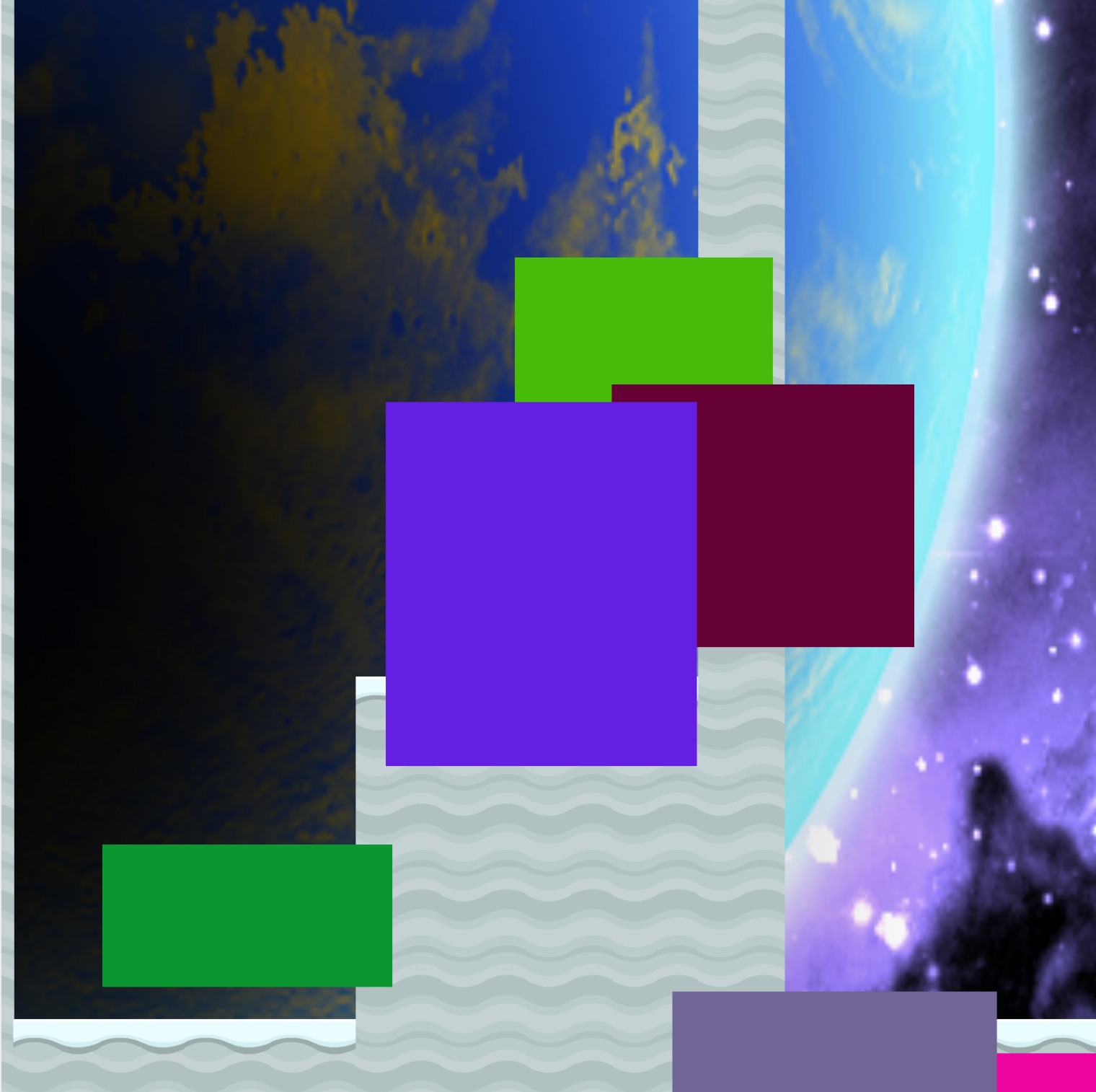}
\end{minipage}%
\hspace*{\fill}
\begin{minipage}{.475 \textwidth}
  \centering
  \includegraphics[width=\textwidth]{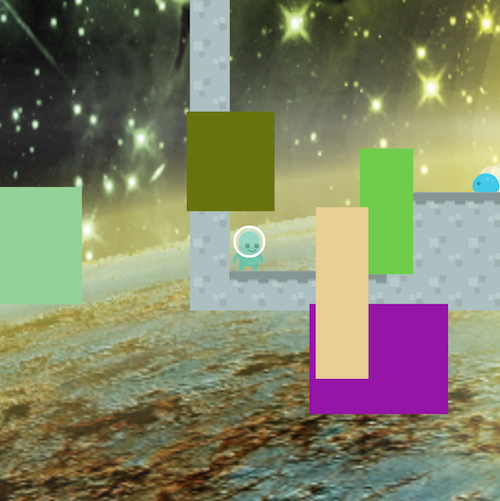}
\end{minipage}
\end{figure}

\section{Hyperparameters and Settings} \label{appendix:hyperparameters}

We used the following hyperparameters and settings in our baseline experiments with the 3 environments. Notably, we forgo the LSTM in CoinRun, and we use more environments per worker in CoinRun-Platforms.

\begin{center}
 \begin{tabular}{||c c c c||} 
 \hline
  & CoinRun & CoinRun-Platforms & RandomMazes \\ [0.5ex] 
 \hline\hline
 $\gamma$ & .999 &  .999 & .999 \\ 
 \hline
 $\lambda$ & .95 & .95 & .95 \\
 \hline
 \# timesteps per rollout & 256 & 256 & 256 \\
 \hline
 Epochs per rollout & 3 & 3 & 3 \\
 \hline
 \# minibatches per epoch & 8 & 8 & 8 \\ 
 \hline
 Entropy bonus ($k_H$) & .01 & .01 & .01 \\
 \hline
 Adam learning rate & \num{5e-4} & \num{5e-4} & \num{5e-4} \\
 \hline
 \# environments per worker & 32 & 96 & 32 \\ 
 \hline
 \# workers & 8 & 8 & 8 \\
 \hline
 LSTM? & No & Yes & Yes \\
 \hline
\end{tabular}
\end{center}

\clearpage
\section{Performance} \label{appendix:performance}

\begin{table}[!h]
\small
\begin{center}
 \begin{tabular}{||c c c c c||} 
 \hline
 \# Levels & Nature Train & Nature Test & IMPALA Train & IMPALA Test \\ [0.5ex] 
 \hline\hline
  100 & $99.45 \pm 0.09$ & $66.79 \pm 1.09$ & $99.39 \pm 0.08$ & $66.58 \pm 1.91$ \\ 
  \hline
  500 & $97.85 \pm 0.46$ & $70.54 \pm 0.62$ & $99.16 \pm 0.19$ & $80.25 \pm 1.07$ \\ 
  \hline
  1000 & $95.7 \pm 0.65$ & $72.51 \pm 0.68$ & $97.71 \pm 1.04$ & $84.84 \pm 2.24$ \\ 
  \hline
  2000 & $92.65 \pm 0.71$ & $75.6 \pm 0.28$ & $97.82 \pm 0.32$ & $90.92 \pm 0.45$ \\ 
  \hline
  4000 & $90.18 \pm 1.04$ & $78.35 \pm 1.47$ & $97.7 \pm 0.19$ & $95.87 \pm 0.62$ \\ 
  \hline
  8000 & $88.94 \pm 1.08$ & $84.02 \pm 0.96$ & $98.13 \pm 0.21$ & $97.29 \pm 1.04$ \\ 
  \hline
  12000 & $89.11 \pm 0.58$ & $86.41 \pm 0.46$ & $98.14 \pm 0.56$ & $97.51 \pm 0.46$ \\ 
  \hline
  16000 & $89.24 \pm 0.77$ & $87.58 \pm 0.79$ & $98.04 \pm 0.17$ & $97.77 \pm 0.68$ \\
  \hline
  $\infty$ & $90.87 \pm 0.53$ & $90.04 \pm 0.9$ & $98.11 \pm 0.26$ & $98.29 \pm 0.16$ \\ 
  \hline
\end{tabular}
\caption{\label{tab:coinrun_gen}CoinRun (across 5 seeds)}
\end{center}
\end{table}

\begin{table}[!h]
\small
\begin{center}
 \begin{tabular}{||c c c||} 
 \hline
 \# Levels & Train & Test \\ [0.5ex] 
 \hline\hline
  100 & $14.04 \pm 0.86$ & $2.22 \pm 0.2$ \\ 
  \hline
  400 & $12.16 \pm 0.11$ & $5.74 \pm 0.22$ \\ 
  \hline
  1600 & $10.36 \pm 0.35$ & $8.91 \pm 0.28$ \\ 
  \hline
  6400 & $11.71 \pm 0.73$ & $11.38 \pm 0.55$ \\ 
  \hline
  25600 & $12.23 \pm 0.49$ & $12.21 \pm 0.64$ \\ 
  \hline
  102400 & $13.84 \pm 0.82$ & $13.75 \pm 0.91$ \\ 
  \hline
  409600 & $15.15 \pm 1.01$ & $15.18 \pm 0.98$ \\ 
  \hline
  $\infty$ & $15.96 \pm 0.37$ & $15.89 \pm 0.47$ \\ 
  \hline
\end{tabular}
\caption{\label{tab:plat_gen}CoinRun-Platforms IMPALA (across 3 seeds)}
\end{center}
\end{table}

\begin{table}[!h]
\small
\begin{center}
 \begin{tabular}{||c c c||} 
 \hline
 \# Levels & Train & Test \\ [0.5ex] 
 \hline\hline
  1000 & $92.09 \pm 0.35$ & $68.13 \pm 1.3$ \\ 
  \hline
  2000 & $93.57 \pm 2.06$ & $74.08 \pm 1.25$ \\ 
  \hline
  4000 & $92.94 \pm 1.47$ & $77.79 \pm 1.87$ \\ 
  \hline
  8000 & $98.47 \pm 0.4$ & $83.33 \pm 0.66$ \\ 
  \hline
  16000 & $99.19 \pm 0.19$ & $87.49 \pm 1.45$ \\ 
  \hline
  32000 & $98.62 \pm 0.37$ & $93.07 \pm 1.0$ \\ 
  \hline
  64000 & $98.64 \pm 0.14$ & $97.71 \pm 0.24$ \\ 
  \hline
  128000 & $99.06 \pm 0.21$ & $99.1 \pm 0.34$ \\ 
  \hline
  256000 & $98.97 \pm 0.18$ & $99.21 \pm 0.11$ \\ 
  \hline
  $\infty$ & $98.83 \pm 0.71$ & $99.36 \pm 0.16$ \\ 
  \hline
\end{tabular}
\caption{\label{tab:maze_gen}RandomMazes IMPALA (across 3 seeds)}
\end{center}
\end{table}

\end{document}